\newif\ifdraft
\draftfalse

\documentclass[a4paper,fleqn]{cas-sc}

\usepackage[numbers]{natbib}

\usepackage[utf8]{inputenc}
\usepackage[T1]{fontenc}
\usepackage[english]{babel}
\usepackage{amsmath, amssymb, amsfonts, amsthm}
\usepackage{graphicx}
\usepackage{fancyhdr}
\usepackage{color}
\usepackage{hyperref}
\usepackage{bbm}
\usepackage{graphicx}
\usepackage{graphbox}
\usepackage[capitalize]{cleveref}
\Crefname{equation}{Eq.}{Eqs.}
\Crefname{figure}{Fig.}{Figs.}
\Crefname{tabular}{Tab.}{Tabs.}
\Crefname{appendix}{Appendix}{Appendices}
\Crefname{section}{Sec.}{Secs.}

\usepackage{array}
\usepackage{tabularx}

\usepackage{bbold}			

\newcommand{\pd}{\textit{PredDiff}}

\usepackage[normalem]{ulem} 
\newcommand{\stkout}[1]{\ifmmode\text{\sout{\ensuremath{#1}}}\else\sout{#1}\fi}

\ifdraft

\newcommand{\deleted}[1]{\textcolor{red}{\stkout{#1}}}

\newcommand{\deletedfloat}[1]{}
\newcommand{\commented}[1]{\textcolor{blue}{#1}}
\else

\newcommand{\deleted}[1]{}

\newcommand{\deletedfloat}[1]{}
\newcommand{\commented}[1]{}
\fi

\begin{document}
\let\WriteBookmarks\relax
\def\floatpagepagefraction{1}
\def\textpagefraction{.001}
\shorttitle{PredDiff: Explanations and Interactions from Conditional Expectations}
\shortauthors{Bl\"ucher et~al.}


\title [mode = title]{PredDiff: Explanations and Interactions from Conditional Expectations}                      

\author[1]{Stefan Bl\"ucher}[orcid=0000-0002-6330-7996]
\fnmark[1]
\ead{bluecher@tu-berlin.de}
\credit{Methodology, Software}

\author[2]{Johanna Vielhaben}[orcid=0000-0001-9399-5710]
\ead{johanna.vielhaben@hhi.fraunhofer.de}
\credit{Methodology, Software}
\author[3]{Nils Strodthoff}[orcid=0000-0003-4447-0162]
\cormark[1]
\fnmark[1]
\ead{nils.strodthoff@uol.de}
\cortext[cor1]{Corresponding author}
\address[1]{Machine Learning Group, TU Berlin, Marchstr. 23, 10587 Berlin, Germany}
\address[2]{Applied Machine Learning Group, Fraunhofer Heinrich-Hertz-Institut, Berlin, Germany}
\address[3]{Division AI4Health, Oldenburg University, Oldenburg, Germany}
\credit{Conceptualization of this study, Methodology, Software}

\fntext[fn1]{Equal contribution}
\fntext[asd]{This is the accepted version of the manuscript that is now published under \href{https://doi.org/10.1016/j.artint.2022.103774}{doi: 10.1016/j.artint.2022.103774}}
\begin{abstract}
		PredDiff is a model-agnostic, local attribution method that is firmly rooted in probability theory. Its simple intuition is to measure prediction changes while marginalizing features. In this work, we clarify properties of PredDiff and its close connection to Shapley values. We stress important differences between classification and regression, which require a specific treatment within both formalisms. We extend PredDiff by introducing a new, well-founded measure for interaction effects between arbitrary feature subsets. The study of interaction effects represents an inevitable step towards a comprehensive understanding of black-box models and is particularly important for science applications. Equipped with our novel interaction measure, PredDiff is a promising model-agnostic approach for obtaining reliable, numerically inexpensive and theoretically sound attributions.
\end{abstract}



\maketitle

\section{Introduction}
Understanding complex machine learning models is fundamental for high-stake applications, e.g., in healthcare or criminal justice. To this end, the Explainable AI (XAI) community has put forward a plethora of different attribution methods, see \cite{covert2021explaining,Lapuschkin_2019,lundberg2017unified,Montavon2018,Samek2021} for reviews.
Most methods summarize the complex, non-linear interactions that a single feature undergoes while traversing a machine learning model into a single attribution score. 
While this approach can provide invaluable informative heatmaps, feature-wise relevances do not provide access to feature interactions \cite{Toothaker1994MultipleRT,tsang2021interpretable} and can even be misleading as interaction effects are implicitly distributed onto single-feature relevances \cite{deng2020unified}.

We envision various applications where the understanding of interaction effects is instrumental to extract knowledge about underlying mechanisms from a machine learning model. We exemplify the prospects for such methods in two domains: firstly, in natural sciences and secondly, in healthcare. 
In the first case, consider a model that is trained to infer protein-protein interactions framed as a binary classification task given both primary protein sequences as input.
Interpretability methods that allow quantifying interaction effects would then enable to identify corresponding binding sites in both sequences.
In the second case, we consider a  medical risk prediction model, which infers the mortality risk based on multiple demographic features and lab values. Here, relying only on single-feature importance might lead to a misleadingly simple picture, as multiple risk factors interact and hence, aggravate or alleviate the mortality risk (such as age and sex in the simplest case). Thus, interaction measures are necessary to capture the complex underlying physiological reality.

In this work, we revisit Prediction Difference analysis (\pd{}), which was originally introduced in \cite{Sikonja2008Explaining}.
In our opinion, the beauty of \pd{} lies in its simplicity and strong connection to probability theory. 
The whole formalism is fixed by marginalizing variables and measuring prediction differences. 
It has been successfully applied on various image classification tasks \cite{gu2020contextual,Tian2017,Wei2018Explain,zintgraf2017visualizing} and also in Natural Language Processing, where it is referred to as input marginalization \cite{harbecke2020considering,kim2020interpretation}.
However, all previous studies miss a comprehensive treatment in a well-controlled setting, testing analytical and experimental limits of \pd{}.
The unifying perspective on perturbation-based attribution methods in \cite{covert2021explaining} shows how \pd{} is closely connected to Shapley values \cite{lundberg2017unified,strumbelj2013explaining} and other approaches of this category.
In particular, \pd{} encompasses single-shot attribution methods, such as occlusion \cite{Zeiler2014} or inpainting image parts with generative models~\cite{agarwal2020explaining,lenis2020domain}. These are, however, not covered by the foundations of \pd{} and potentially unreliable.

Our main contribution is a novel interaction measure for \pd{}. It is well-founded and allows decomposing feature relevances into main and joint effects. 
Importantly, our decomposition is applicable to any interaction order and obeys a completeness relation. 
Incorporating and quantifying feature interactions has very recently attracted interest within the XAI community, see \cite{tsang2021interpretable} for a review.
First works on interaction effects appeared in \cite{deng2020unified,Eberle_2020,janizek_explaining_2020,schnake_higher-order_2020,Tian2017}.
Additionally, interaction measures have been proposed for Shapley value-based approaches \cite{gosiewska2020trust,lundberg2020local,pmlr-v119-sundararajan20a,zhang2021interpreting}, global ALE-plots \cite{apley2016visualizing}, and other perturbation-based approaches \cite{Jin2020Towards,tsang2020does}.
\pd{} has the particular advantage that it allows quantifying interaction effects for arbitrary (non-overlapping) feature sets, while remaining an optimal linear scaling.

Additionally, we investigate \pd 's theoretical properties and demonstrate its intimate relationship to Shapley values.
In particular, we shed light on the intricacies of classification due to the inherent connection between the classifier and the underlying data distribution.
We present consequences for \pd{} and Shapley values, both on the level of relevances and interactions.
Finally, we present experimental evidence for the soundness of our framework and find qualitative agreement to the popular Shapley Interaction Index. In particular, we quantify feature interactions for an image classifier, as a task that is already intractable for many competing methods with a less favorable scaling than \pd{}.

To summarize, our main contributions are 
(i) an investigation of the theoretical properties of \pd{} and its relation to Shapley values 
(ii) a novel interaction measure based on a proper functional decomposition, satisfying an \emph{interaction completeness} property on relevance level
(iii) an analysis of intricacies of classification due to the inherent connection between classifier and data distribution
(iv) the experimental validation on analytic, synthetic and real-world datasets for classification and regression.

\section{\pd: a local, model-agnostic, probabilistically sound attribution method}\
We specify our notation as follows. $\mathcal{X}=\{X^1, \ldots, X^{N}\}$ is our set of $N$ features. Uppercase letters ($X^a$) denote the features itself (with unspecified values) and lowercase letters ($x^a$) refer to a specific instance. Additionally, we routinely split all features into pairwise disjoint subsets $X, Y$ and $Z$ with $\mathcal{X}=X\cup Y \cup Z$. Typically, we assess the interaction relevance between feature sets $Y$ and $Z$ in the presence of the remaining set of features $X$.

\subsection{Relevances for classification and regression tasks}
\label{sec:pdrelevances}
We consider a classification task, where a classifier $f_c$ provides access to the conditional probability of class $c$, i.e., $f_c(x,y):=p(c|x,y)$.
One way of assessing the relevance of a particular set of features $Y$ is to compare the original prediction $p(c|x,y)$ to the prediction $p(c|x)$, where the feature(s) $Y$ has(ve) been removed.
For an arbitrary classifier, this can be implemented in a probabilistically sound manner by \emph{m-}arginalizing $Y$ \cite{Sikonja2008Explaining} via
\begin{equation}
	\label{eq:defmclas}
	m^{f_c}_{Y|x} := p(c|x)  = \int p(c|x,Y) p(Y|x) \text{d}Y \approx \int f_c(x,Y) q(Y|x) \text{d}Y \,.
\end{equation}
Here, $p(Y|x)$ represents the true generative distribution for reconstructing $Y$ given the remaining features $X$ evaluated at $x$. In practice, we typically draw a fixed number of random samples from an empirical imputer distribution $q(Y|x)$ that approximates $p(Y|x)$, see \Cref{app-sec:sampleapproximation} for more numerical details. Therefore, \pd{} does not suffer from an unfavorable factorial scaling with the number of involved features.
Additionally, one straightforwardly obtains confidence intervals for relevance scores via empirical bootstrapping.
In this sense, the approach is completely domain- and task-agnostic, provided an appropriate generative model for imputation. In terms of imputer distributions $q(Y|x)$, one can broadly distinguish between \emph{marginal imputer distributions}, which completely neglect the dependence on $x$ and therefore in general inevitably produce off-manifold samples, and $x$-dependent \emph{conditional imputer distributions}. It is worth noting that all perturbation-based attributions methods have to deal with this issue. For Shapley values, this is captured in the recent discussion on interventional as compared to observational Shapley values \cite{pmlr-v108-janzing20a,kumar2020problems,sundararajan2020shapley}. In our experiments, we always present results for a conditional as well as a marginal imputer to give the reader a qualitative impression of the impact of imputer choice. A detailed comparison is deferred to future work. As a final remark, we stress that the probabilistic interpretation of \Cref{eq:defmclas} clearly requires the use of a conditional imputer distribution.

In general, \pd{} relevances are obtained by comparing the occluded prediction to the sample prediction. 
Several possibilities have been proposed in the literature \cite{Sikonja2008Explaining}. 
Here, we compare logarithmic differences, which are interpreted as the information difference conveyed by $Y$, i.e.,\
\begin{equation}
	\label{eq:defmhat}
	\bar m^{f_c}_{Y|x} := \log_2 f_c(x, y) - \log_2 m^{f_c}_{Y|x}\,,			
\end{equation}
see \Cref{sec:interaction} for a novel argument favoring this choice. We stress that this equally applies to other attribution methods, see \Cref{sec:shapley}. We avoid issues with vanishing probabilities, as in \cite{Sikonja2008Explaining}, by means of a Laplace correction, i.e.,\ by mapping $p \to (p \,M + 1)/(M + K)$, where $M$ is the number of training instances and $K$ the number of classes.
As a second remark, the probabilistic interpretation of \Cref{eq:defmclas} relies on the identification of $f_c$ with a proper probability distribution. On general grounds, having a \textit{well-calibrated classifier} is desirable for any classifier. 
In this work, we use temperature-scaling, which scales the pre-softmax activations by a single global scaling factor in order to shift the prediction confidence appropriately, to achieve this \cite{Guo2017Calibration}.

Turning to regression problems, where we infer relevances with respect to a particular model $f(x,y)$.
Hence, only the class subscript  in \Cref{eq:defmclas} is suppressed, i.e.,\
\begin{equation}
	\label{eq:defmreg}
	m^{f}_{Y|x} := \int f(x, Y) p(Y|x) \text{d}Y \,.
\end{equation}
Here, the target is directly meaningful and therefore, we directly consider centered $m$-values via \cite{letzgus2021toward,lundberg2017unified,strumbelj2013explaining}
\begin{equation}
	\label{eq:defmbar}
	\bar m^{f}_{Y|x} :=  f(x,y) - m^{f}_{Y|x}\,, 
\end{equation}
with a slight abuse of notation in order to unify regression and classification tasks as far as possible.

We discuss different properties of \pd{} attributions in \Cref{app-sec:preddiff_properties}. \pd{} satisfies the classic Shapley axioms on \emph{sensitivity}, \emph{linearity} and \emph{symmetry}.
However, the \emph{completeness axiom}--i.e., summing up all feature relevances is equal to the prediction plus some reference value--only holds under particular circumstances.
However, it holds in the case where it is indispensable, namely for linear models with independent features. Here, \pd{} relevances in fact coincide with Shapley values, see \Cref{app-sec:linear}. 
We present a comprehensive discussion of the \emph{completeness axiom} in \Cref{sec:efficiency}.

\subsection{Interaction relevances}
\label{sec:interaction}

\subsubsection{Decomposition and completeness relation}
We start in a regression setting, which is conceptually slightly simpler.
The intuition behind our approach is to decompose the model prediction $f$ into its main, additive  components $f^Y$/$f^Z$ and its interactive, non-additive part $f^{YZ}$. Subsequently, this induces a similar decomposition on the level of relevances, which we use to measure interactions effects between the features $Y$ and $Z$. 
This is achieved by using the anchored expansion from \cite{kuo2010decompositions} with the sample $(x,y,z)$ as anchor point. This results in a decomposition of the form (and already evaluated at $X=x$),
\begin{equation}
	\label{eq:decompositionxyz}
	f(x,Y,Z) = f^{\varnothing} + f^{Y}(Y) +f^{Z}(Z) + f^{Y\!Z}(Y,Z)\,,
\end{equation}
where $f^\varnothing = f(x,y,z)$, $f^Y(Y)=f(x,Y,z)-f(x,y,z)$, $f^Z(Z)=f(x,y,Z)-f(x,y,z)$ and $f^{Y\!Z}(Y,Z)=f(x,Y,Z)-f(x,Y,z)-f(x,y,Z)+f(x,y,z)$, see \Cref{app-sec:two_variable_decomposition} for details. Here, the superscripts denote the remaining functional dependence, e.g., $f^{Y}$ is only a function of $Y$.
The decomposition is unique in the sense that it is the only decomposition that fulfills the annihilation property, i.e., $f^\alpha=0$ if any feature set in $\alpha$ is set to its anchor point value. The decomposition is minimal in the sense that it avoids unnecessary higher-order terms as far as possible \cite{kuo2010decompositions}, which is a desired property in our case \cite{yu2019reluctant}.
Now, we can use \Cref{eq:defmbar} to compute \pd{} relevances for \Cref{eq:decompositionxyz} and obtain a \emph{completeness relation}, which constitutes the heart of our formalism, i.e.,
\begin{equation} \label{eq:rel_sum}
	\bar{m}^f_{Y\!Z|x} = \bar{m}^{f^Y}_{Y|x} + \bar{m}^{f^Z}_{Z|x} +\bar{m}^{f^{Y\!Z}}_{Y\!Z|x}\,,
\end{equation}
where we used that $\bar{m}^{f^\varnothing}_{\varnothing|x}=0$ and $\bar{m}^{f^Y}_{Y|x}=\bar{m}^{f^Y}_{Y\!Z|x}$ (as by definition $f^Y$ does not depend on $Z$). The interpretation of the different terms will be discussed in \Cref{sec:interpretation}. 
Using \Cref{eq:defmbar}, the quantity of interest $\bar{m}^{f^{Y\!Z}}_{Y\!Z|x}$ is thus explicitly given by
\begin{equation}
	\begin{split}
	\bar{m}^{f^{Y\!Z}}_{Y\!Z|x} &= - m^f_{Y\!Z|x} + m^{f^Y}_{Y|x} + m^{f^Z}_{Z|x} + f(x,y,z)\\
	&=-\int f(x,Y,Z) p(Y,Z|x) \text{d}Y \text{d}Z + \int f(x,Y,z) p(Y|x) \text{d}Y + \int f(x,y,Z) p(Z|x) \text{d}Z - f(x,y,z) \,.
	\end{split}
\end{equation}
In particular, $\bar{m}^{f^{Y\!Z}}_{Y\!Z|x}$ vanishes in the case of a non-interacting regressor of the form $f(X,Y,Z) = h(X,Y) + g(X,Z)$. We refer to this property as the \emph{no-interaction} property. As an important remark, the different constituents on the right-hand-side of \Cref{eq:rel_sum} inherit the computation complexity of the original \pd{} relevances. Anchoring the decomposition at the sample point $(x,y,z)$ is the only consistent choice within the \pd{} framework, see \Cref{app-sec:anchorpoint} for a detailed discussion.
Generalizing the decomposition \Cref{eq:decompositionxyz} to an arbitrary number of interacting feature sets, i.e., higher-order effects, is straightforward and leads to the \emph{interaction completeness} property \Cref{app-eq:completeness_threepoint}, which is analogous to \Cref{eq:rel_sum}, see \Cref{app-sec:npoint_interaction}~and~\ref{app-sec:npoint_completeness_relation} for details.

\subsubsection{Interpretation}
\label{sec:interpretation}

We now work out an interpretation for the individual terms of the \emph{interaction completeness} \Cref{eq:rel_sum}. The left-hand-side of the equation relates to the prediction change, i.e., loss or gain of information, when both feature sets $Y, Z$ are occluded. 
Therefore, the interpretation of the raw \pd{} effects follows to be:
\begin{itemize}
	\item \emph{(raw) main effect} $\bar{m}^{f^Y}_{Y|x}$: Prediction difference corresponding to solely occluding $Y$ with knowledge of all other features. Hence, contains all higher-order joint effects at fixed values of the interaction partners $Z=z$.
	\item \emph{(raw) joint effect} $\bar{m}^{f^{Y\!Z}}_{Y\!Z|x}$: Interactive prediction difference corresponding to jointly occluding $Y$ and $Z$ in a way that is not covered by a single corresponding main effect, i.e., by keeping either feature fixed at $y$ or $z$, respectively.
	
\end{itemize}
Finally, we refer to $\bar{m}^{f}_{Y|xz}$ as \emph{(raw) relevances}, which agree with the corresponding main effect $\bar{m}^{f^Y}_{Y|x}$ up to the used conditioning.
For regression, we can additionally define \emph{shielded} counterparts, which specifically exclude the combined feature effects from the main effect.
This point of view relies on regrouping terms in \Cref{eq:rel_sum} and leads to an alternative decomposition of $\bar{m}^f_{Y\!Z|x}$ of the form
\begin{align}
	\label{eq:completenessshielded}
	\bar{m}^f_{Y\!Z|x} &= \bar{m}^{f^{Y}}_{Y\backslash Z|x} + \bar{m}^{f^{Z}}_{Z\backslash Y|x} + \bar{m}^{f^{Y\!Z}}_{\backslash Y\!Z|x}\,,
\end{align}
where $\bar{m}^{f^Y}_{Y\backslash Z|x} :=  \bar{m}^{f^{Y\!Z}}_{Y\!Z|x} + \bar{m}^{f^Y}_{Y|x}=\bar{m}^{f(x, Y, Z)}_{Y\!Z|x} - \bar{m}^{f(x, y, Z)}_{Z|x}$ and $\bar{m}^{f^{Y\!Z}}_{\backslash Y\!Z|x}:=-\bar{m}^{f^{Y\!Z}}_{Y\!Z|x}$.

The terms in \Cref{eq:completenessshielded} have the following interpretation:
\begin{itemize}
	\item \emph{shielded main effect} $\bar{m}^{f^Y}_{Y\backslash Z|x}$: Prediction difference corresponding to solely occluding $Y$ without the presence of $Z$. Hence, it is \emph{shielded} from the joint effect between $Y$ and $Z$. 
	\item \emph{shielded joint effect} $\bar{m}^{f^{Y\!Z}}_{\backslash Y\!Z|x}$: Interactive prediction difference corresponding to jointly occluding $Y$ and $Z$, i.e., the super-additive part with respect to the shielded main effects.
\end{itemize}
We show how shielded effects can be constructed for third order interactions in \Cref{app-sec:shielded3pt}.
To build a better intuition, note that, under the assumption of a factorizing imputer distribution $p(y, z|x)=p(y|x)p(z|x)$, we can write
\begin{equation}
	\bar{m}^{f^{Y}}_{Y\backslash Z|x}= f^{\backslash Z}(x, y) - \int \text{d}Y f^{\backslash Z}(x,Y) p(Y|x) = \bar{m}^{f^{\backslash Z}}_{Y|x}\,,\\
\end{equation}
where $f^{\backslash Z}(x, y) = \int \text{d}Z f(x, y, Z)p(Z|x)$. The shielded main effect is, therefore, nothing but the main effect of the model where $Z$ has been marginalized.

\subsubsection{Classification}
\label{sec:classification}
For classification settings, the situation is more intricate due to the fact that the model's class-conditional probabilities and the data distribution are implicitly tied, as both relate to the joint distribution of labels and input features. We rely on \pd{} relevances for classification as additive information differences and postulate that the completeness relation \Cref{eq:rel_sum} remains valid in the classification setting, i.e., upon replacing $f$ by $f_c$. We support this argument by investigating the
\emph{no-interaction} property, as a necessary condition for any sensible interaction measure, which entails that a non-interacting classifier yields a vanishing interaction relevance. To define a non-interacting classifier, we consider a generalization of \textit{informative conditional interactions} \cite{henelius2017interpreting,jakulin2003quantifying}, which implies that there is no label $c$ and residual features $X$ such that the feature sets $Y$ and $Z$ interact directly. Thus, we define a classifier where $Y$ and $Z$ do not interact by
\begin{equation}
	\label{eq:defnointeraction}
	p(Y,Z|c, x)= p(Y|c, x) p(Z|c, x)\,.
\end{equation}
If one works out \pd{} relevances using this assumption, see \Cref{app-sec:derivationinteraction}, one is lead to the joint effect
\begin{equation}
	\label{eq:classificationmi}
	\begin{split}
		\bar{m}^{f_c^{Y\!Z}}_{Y\!Z|x}&=\bar{m}^{f_c}_{Y\!Z|x}-\bar{m}^{f^Y_c}_{Y\!Z|x}-\bar{m}^{f^Z_c}_{Y\!Z|x}\\
		&= \log_2\left(\frac{p(y,z|x)}{p(z|x)p(y|x)}\right) + \log_2\left(\frac{1}{p(x|c)}\int \text{d}Y\text{d}Z p(x, Y, Z|c)\frac{p(Y|x)}{p(Y|x,z)}\frac{p(Z|x)}{p(Z|x,y)}\right) \,.	
	\end{split}
\end{equation}
where the first term is conventionally referred to as \textit{local conditional mutual information}. The second term relates to the conditioning, i.e., to using $\bar{m}^{f^Y_c}_{Y\!Z|x}=\bar{m}^{f^Y_c}_{Y|x}$ instead of $\bar{m}^{f}_{Y|xz}$, which is inevitable if one insists on comparing only objects that share a common conditioning, as it was done in the regression case, see \Cref{app-sec:derivationinteraction} for details. 
Also the occurrence of the first term is naturally explained by the fact that classifier and data distribution are tied (through a constraint on the joint distribution \Cref{eq:defnointeraction}) in the sense that the information difference on the left-hand-side also yields a term that just quantifies the information difference on the level of the input features. 
These terms are not specific to \pd{} but naturally appear also in other formalisms such as Shapley values in a classification setting, see \Cref{sec:shapley}. Lastly, it is worth stressing that the \emph{no-interaction} property singles out logarithmic differences in \Cref{eq:defmhat} and does not hold for other popular difference measures, such as raw probabilities or log-odds \cite{Sikonja2008Explaining}, see \Cref{app-sec:derivationinteraction}.

At this point, there are different ways to ensure $\bar{m}^{f_c^{Y\!Z}}_{Y\!Z|x}=0$ for a non-interacting classifier as defined by \Cref{eq:defnointeraction}. For conventional discriminative models, one would use a separate generative model (imputer) $q(Y,Z|x)$ to approximately sample from $p(Y,Z|x)$, which unties the relation between the output probabilities and the data distribution.
Here, we proceed by noting that both terms on the right-hand-side vanish upon using a factorizing imputer distribution $q(Y,Z|x) = q(Y|x)q(Z|x)$, which also implies $q(Y|x)=q(Y|x,z)$ and $q(Z|x)=q(Z|x,y)$.
This can be implemented by sampling two copies $(y_1,z_1)$ and $(y_2,z_2)$ from $p(Y,Z|x)$ and using $(y_1,z_2)$ and $(y_2,z_1)$, see \Cref{app-sec:sampleapproximation} for details.
Firstly, this exposes the classifier to samples that are off-manifold to a slight degree, as the connection between $Y$ and $Z$ has been broken, and secondly, induces a sampling error due to the fact the sampling distribution $q(Y|x)q(Z|x)$ does not capture the implicit relation between $Y$ and $Z$ in $p(Y,Z|x)$.
We see this as a minor issue as this sampling error will most likely still be smaller than the inherent approximation error arising from training the imputer $q(Y,Z|x)$ to match $p(Y,Z|x)$ based on a limited amount of data. Alternatively, for hybrid models that provide access to the joint probability $p(x,y,z)$, such as \cite{Grathwohl2020YourCI}, or imputer that provide an exact sampling probability (notwithstanding the inevitable mismatch between imputer and data distribution), such as normalizing flows \cite{Kobyzev2021}, one option would be to compute the terms on the right-hand-side and to subtract them from the left-hand-side in order to define the joint effect.

\subsection{Implications for Shapley values}
\label{sec:shapley}

\subsubsection{Connection between Shapley and \pd{}}
\label{sec:shapleypdconnection}
Shapley values are a popular tool for local model-agnostic attribution \cite{lundberg2017unified,strumbelj2013explaining} based on game theory \cite{shapley1953value}.
In general, Shapley values are given by
\begin{equation}
	\phi_{j}(v)=\sum_{S \subseteq\mathcal{X} \backslash\left\{X^{j}\right\}} \frac{|S| !(N-|S|-1) !}{N !}
	\left[v(S \cup X^j) - v(S)\right]\,.
	\label{eq:defshapley}
\end{equation}
The remaining ambiguity is to specify a connection between a model $f$, an instance $x$ and the value function $v(S)$. A common choice uses an observational (conditional) distribution to occlude redundant features $X_{\bar{S}}\notin S$, i.e.,
\begin{equation}
	v^\text{reg}_{f, x}(S) = \mathbb{E}_{X_{\bar{S}}|x_S}\left[f(x_S, X_{\bar{S}})\right] = m^{f}_{X_{\bar{S}}|x_S}.
	\label{eq:defvaluefct_regression}
\end{equation}
One then identifies the first Shapley term $S=\mathcal{X} \backslash\left\{X^{j}\right\}$ with the \pd{} relevance $\bar{m}^f_{X^j|x_S}$. This reveals an intimate connection between both formalisms. 
However, there is an ongoing debate whether one should replace the observational by an interventional (marginal) distribution, see \cite{pmlr-v108-janzing20a,kumar2020problems,sundararajan2020shapley}. This would break the previous correspondence. In general, marginal distributions generate illegitimate, out-of-distributions samples, questioning the reliability of resulting attributions. Additionally, ignoring feature dependencies unavoidably leads to simple adversarial attack strategies \cite{anders2020fairwashing,Slack2020}.

Turning to feature interactions, we consider the relation to the 'Shapley Interaction Index' \cite{lundberg2020local}, which was proposed as an explicit measure for interactions based on game theory \cite{Fujimoto2006}.
Interestingly, we can map \pd 's shielded joint effect onto their central object, a discrete second order derivative, i.e.,
\begin{align}
	\bar{m}^{f^{Y\!Z}}_{\backslash Y\!Z|x} = f(x, y, z) -  m^{f(x, Y, z)}_{Y|x} - m^{f(x, y, Z )}_{Z|x} + m^{f(x, Y, Z)}_{Y\!Z|x} = \delta_{YZ}(S)\,,
	\label{eq:shapleyinteractionindex}
\end{align}
for $S=\mathcal{X}\backslash{YZ}$ and $Y$/$Z$ restricted to a single feature each.
In the same setting, \pd 's shielded main effects, such as $\bar{m}^{f^{Y}}_{Y \backslash Z|x}$, can be identified with the second Shapley term. This reiterates the close connection between both formalisms, which we expect to hold at higher orders as well. 
A different proposed interaction measure within the Shapley value formalism is the 'Shapely Taylor Interaction Index' \cite{pmlr-v119-sundararajan20a}. 
It is centered around a general discrete derivative formula, which allows incorporating arbitrary interaction orders. 
Here, we point out that these discrete derivatives are identical to the general decomposition underlying \Cref{eq:decompositionxyz} up to a global sign, see \Cref{app-sec:two_variable_decomposition} for details.

\subsubsection{Common challenges for classification} \label{sec:classification_shapley}

Here, we leverage our insights from the \pd{} discussion on classification in \Cref{sec:classification} and revisit the foundations of Shapley values within a classification setting. To the best of our knowledge, this topic has so far scarcely received attention in the literature and there is no rigorous argument for either measure, see \cite{strumbelj2013explaining} for possible choices. In the following, we introduce a novel argument based on the \emph{no-interaction} property, which clearly favors logarithmic $m$-values.

As already stated previously, there is no fundamental rule connecting a classifier $p(c|\mathcal{X})$ to the Shapley value function $v_{f_c, x}(S)$.
As for \pd{},  the occluded raw probabilities $p(c|x_S)$ are the most natural object to base the value function on. Drawing further inspiration from \pd{}, we propose to use \emph{logarithmic m-values}, i.e.,
\begin{equation} \label{eq:defvaluefct_classification}
	v_{f_c, x}(S) = \log_2\left(p(c|x_S)\right) = \log_2 \mathbb{E}_{X_{\bar{S}}|x_S} p(c|X_{\bar{S}},x_S)= \log_2\left(m^{f_c}_{X_{\bar{S}}|x_S}\right).
\end{equation}

To demonstrate the benefits of this choice, consider a non-interacting classifier $p(c|\mathcal{X})$ with $\mathcal{X}=\{X^a, X^b\}$ and $p(X^a, X^b|c) = p(X^a|c)p(X^b|c)$ \cite{henelius2017interpreting,jakulin2003quantifying}, as a simplified version of \Cref{eq:defnointeraction}. In this case, the attribution of feature $X^a$ is supposed to not depend on $p(c|x^b)$. Indeed, one easily derives the Shapley value for $X^a$,
\begin{align}
	\phi_a(v) &= \frac{1}{2}\left(v(\{X^a\})-v(\varnothing)+v(\{X^a,X^b\})-v(\{X^b\})\right)\nonumber\\
	&= \log_2 p(c|x^a) - \log_2 p(c) + \frac{1}{2} \log_2 \frac{p(x^a)p(x^b)}{p(x^a,x^b)}\,,\label{eq:shapley_noninteracting}
\end{align}
where we obtained the second line by inserting the definition of the value function \Cref{eq:defvaluefct_classification} and used Bayes' rule in conjunction with the non-interacting classifier Ansatz to write $p(c|x^a,x^b)= p(c|x^a)p(c|x^b)\frac{p(x^a)p(x^b)}{p(c)p(x^a,x^b)}$. Due to occurrence of the logarithm, the second independent classifier $p(c|x^b)$ cancels from the final expression as required. Hence, the Shapley values now independently rely on the respective classifier and the corresponding data distribution.
The last term is inevitable and a consequence that predictors for different classes are inherently tied, see the discussion in \Cref{sec:classification}.

It is worth stressing that such a cancellation does not take place upon using the value function from the regression setting, i.e., $v_{f_c, x}(S) = p(c|x_S)$ with $f_c(x)=p(c|x)$. 
	Here, both independent classifiers interactively define the single feature Shapley values. 
	This value function relates Shapley values to differences of probabilities, see the first line of \Cref{eq:shapley_noninteracting}, which leads to difficulties as for classification the notion of additivity relates to independent--hence factorizing--feature contributions. This clearly invalidates the use of the regression value function in the classification setting. 
As a further remark, other value functions such as $v_{f_c, x}(S) = \mathbb{E}_{X_{\bar{S}}|x_S} \log_2 p(c|X_{\bar{S}},x_S)$, which have been used in related contexts \cite{chen2018learning}, break the natural connection to the occluded raw probabilities but do not resolve this issue. 
We point out that the former value function coincides with \Cref{eq:defvaluefct_classification} if the expectation value is approximated by a single sample, as it is conventionally done for Shapley values.

To summarize, the classification setting poses similar challenges in the Shapley value framework as for \pd{}, even on the level of two single features rather than for entire sets as for \pd{}. The choice of the \pd{} relevance measure in \Cref{eq:defmhat} translates into the choice of the value function for Shapley values.

\subsubsection{Consequences of \emph{no-interaction} properties} \label{sec:shapleyinteraction}
In \Cref{app-sec:shap_interaction} we explicitly evaluate the 'Shapley Interaction Index' w.r.t the \emph{no-interaction} property.
Here, we summarize the main findings:
For both regression and classification, the main issue is that Shapley values are obtained by aggregating attributions obtained from different conditional distributions. 
In the regression setting, one is directly left with differences of conditional distributions. Consequently, the \emph{no-interaction} property can only be satisfied upon using an interventional (marginal) distribution.
For classification, the \emph{no-interaction} property induces an additional constraint on the classifier level, $p(y,z|c)=p(y|c)p(z|c)$, which is in general not satisfied. Thus, the 'Shapley Interaction Index' does not satisfy the \emph{no-interaction} property in the classification setting.

\subsubsection{Comparing \pd{}'s interaction completeness relation vs. Shapley's completeness axiom}
\label{sec:efficiency}
Equipped with insights from \pd{} interaction attributions, including the \emph{interaction completeness} \Cref{eq:rel_sum}, and the intimate relation to Shapley values, it is worthwhile revisiting the Shapley \emph{completeness axiom}, which was already briefly discussed in \Cref{sec:pdrelevances}. 

\paragraph{Off-manifold Evaluation} The \emph{completeness axiom} enforces the Shapley formalism to use the complete set of coalitions $S$. This generally leads to off-manifold evaluations of the underlying predictor. Here, using an interventional distribution leads to maximally off-manifold samples. 
	In principle, this issue could be mitigated through the use of a conditional distribution. However, from a practical point of view, devising high-quality imputers, which produce on-manifold samples upon imputing a large fraction of input variables, remains very challenging, see \Cref{app-sec:cubbirds} for explicit visualizations. Hence, for all practical purposes, this still leads to a certain degree of off-manifold evaluation and consequently unreliable attributions. 
	In this respect, \pd{} takes the least invasive approach as it only requires to impute the variable of interest, i.e., the feature set for which a user wants to compute the relevance for.

\paragraph{Recovering Completeness}
	As a second aspect, we would like to stress that the \emph{completeness axiom} is not lost within the \pd{} framework but recovered after including all interaction effects. Importantly, through explicitly including interaction effects, \pd{} can circumvent potential inconsistencies related to solely considering additive explanations, i.e., see \Cref{subsec:simpleexamples} and \cite{gosiewska2020trust}. 
	This also motivates the phrasing \emph{interaction completeness} property, \Cref{eq:rel_sum} or \Cref{app-eq:completeness_threepoint}, which decomposes the relevance into main effects and higher-order interaction effects. 
	Here, we focus on the case of three variables $A,B,C$ and stress that this argument generalizes to more variables in a straightforward way. In this case the \emph{interaction completeness} \Cref{app-eq:completeness_threepoint} yields {(with $X=\varnothing$)
	\begin{equation}
		\label{eq:3ptcompleteness}
		\bar{m}^f_{ABC}=\bar{m}^{f^A}_A+ \bar{m}^{f^B}_B+ \bar{m}^{f^C}_C+ \bar{m}^{f^{AB}}_{AB}+\bar{m}^{f^{BC}}_{BC}+\bar{m}^{f^{AC}}_{AC}+\bar{m}^{f^{ABC}}_{ABC}=f(a,b,c)-m^f_{ABC}\,,
	\end{equation}
	where $m^f_{ABC}=\int f(A,B,C) p(A,B,C) \text{d}A  \text{d}B \text{d}C$ equals the mean prediction.
	This implies that the sum of all orders \pd{} effects (left-hand-side)  yields the difference between actual and mean prediction (right-hand-side). Hence, the \emph{completeness axiom} is recovered upon including all \pd{} interaction terms. However, evaluating all $\binom{N}{1}+\ldots +\binom{N}{N}=2^N$ terms in the case of $N$ features becomes computationally infeasible for large $N$. This problem is well-known from the Shapley value literature \cite{mitchell2022sampling}.
	 In addition, measuring higher-order interactions is potentially numerically unreliable. The advantage of \pd{} lies in the fact that it allows terminating at a given interaction order. The fact that the right-hand-side of \Cref{eq:3ptcompleteness} agrees with the Shapley result, reiterates that \pd{}, upon including interaction effects, represents a different way of combining Shapley terms $S$ and reveals the close relationship between both formalisms.

\subsection{Favorable computational scaling and practical considerations}
\label{sec:scaling}

In this section, we comment on practical considerations for interaction measures and discuss the superior numerical scaling of \pd{} relevances and joint effects.
In general, the analysis of feature interactions is inherently hindered by the combinatorics of combining all features. 
For $N$ features and a binary interaction measure, this scales as $\mathcal{O}(N^2)$. Here, this issue is circumvented by two effects
\begin{enumerate}
\item Grouping features into semantically meaningful sets, e.g., superpixels obtained from classical segmentation algorithms as in \cite{Wei2018Explain}, which eases this problem significantly.
Further, this renders relevances and interactions based on these feature sets more interpretable. 
This is trivially incorporated in \pd{} and is in principle also possible for Shapley value-based approaches \cite{jullum2021groupshapley}. 
\item Another advantage of \pd{} joint effects is the application on a targeted subset of features. 
We are not bound to evaluate all possible feature combinations but can instead focus on specific features, e.g.,  selecting reference features with high feature relevance and investigating interactions among them and/or with all other features. We demonstrate such an approach in \Cref{subsec:mnist} and \ref{subsec:cub_birds} for image applications.
Alternatively, one could rely on heuristics to group and select interesting combinations of features, as for example done in \cite{Jin2020Towards,tsang2020does}. We propose a procedure within a regression setting in \Cref{subsec:syntheticdataset}. We stress that the previous considerations apply to all interaction measures and are not specific to the \pd{} joint effects.
\end{enumerate}

\begin{table}[bt]
	\caption{Scaling behavior, i.e., required number of model calls $\#$, for different model-agnostic attribution (binary interaction) methods. $N$ feature (sets), selection of $n\leq N$ features ($m \leq N^2$ feature pairs). $l$: number of imputations. }
	\begin{tabular}{p{1.8cm} | p{2.5cm}p{7cm}p{2.8cm}} 
		\multicolumn{1}{c}{}	& 	\pd{}	& 	KernelSHAP \cite{lundberg2017unified}		& 		Shapley values \\ \hline 
		Relevances	&Linear scaling \newline$\mathcal{O}(n l)$	& Solves a global optimization problem. Hence, it computes attributions for all N features simultaneously and therefore, lower bounded by, see \cite[Method 8]{lundberg2020local}, \newline $\mathcal{O}(N l)$.		&	Exponential scaling \newline $\mathcal{O}(N!)$		\\
		(binary) \newline Interactions	&	Linear scaling \newline $\mathcal{O}(m l)$	&	 A hypothetical algorithm would na\"ively require $\mathcal{O}(N^2l)$ model calls, which is unfeasible even for MNIST. 	&	Exponential scaling \newline $\mathcal{O}(N!)$	\end{tabular}
	\label{tab:computationalscaling}
\end{table} 

Next, we compare the computational cost of \pd{} to Shapley value-based approaches, which are the most direct competitors of \pd{}. 
We summarize the different scaling behaviors in \Cref{tab:computationalscaling}. 
Due to the \emph{completeness axiom}, Shapley value-based methods need to correlate all feature attributions. This is either done via sampling feature coalitions or alternatively by dealing with all features simultaneously (KernelSHAP). 
In contrast, \pd{} directly isolates feature attributions and therefore scales optimally with the number of features $N$. Note that the popular occlusion attributions are a one-shot approximation of \pd{}, i.e., use a single model call per feature attribution \cite{agarwal2020explaining,lenis2020domain,Zeiler2014}.
Consequently, \pd{} relevances and interactions achieve the most favorable scaling possible (i.e., only scales with the number of imputations) for model-agnostic, perturbation-based approaches. We explicitly show \pd{}'s computational advantage in \Cref{subsec:mnist}.

\section{Results and discussion}
\subsection{Analytic example}
\label{subsec:simpleexamples}
We revisit a famous example from \cite{strumbelj2010efficient,strumbelj2013explaining}, which has been used as an argument against approaches along the line of \pd{}. 
We consider two binary input variables $X$ and $Y$ that are sampled uniformly, i.e., are subject to the data distribution $p(X,Y)=\frac{1}{4}$. The function under consideration is $f(X,Y)= X \vee Y$. 
For consistency with the literature, we work in a regression setting, but the same qualitative conclusions can be drawn from a classification setting.
\begin{table}[bt]
	\caption{\pd{} raw (left side) and shielded (right side) main and joint effects for $f(X,Y)=X \vee Y$ and a uniform data distribution (up to a constant $^1/_4$). As a consequence of the \emph{interaction completeness} \Cref{eq:rel_sum} and \Cref{eq:completenessshielded}, the column totals on the left side equal those on the right side.}
	\begin{center}
		\begin{tabular}{c|*{4}{c}||c|*{4}{c}}
			\midrule 
			$(x, y)$	\hspace{0pt}&\hspace{0pt}	$(0, 0)$	&	$(1, 0)$	&	$(0, 1)$	&	$(1, 1)$	& $(x, y)$  &	$(0, 0)$	&	$(1, 0)$	&	$(0, 1)$	&	$(1, 1)$	\\ \midrule 
			$\bar{m}^{f^{X}}_{X}$	   &	$-2$	&	$+2$	&	$\phantom{+}0$		&	$\phantom{+}0$	& $\bar{m}^{f^{X}}_{X\backslash y}$ &	$-1$  &	$+1$	&	$-1$ &	$+1$ \\ 
			$\bar{m}^{f^{Y}}_{Y}$	   &	$-2$	&	$\phantom{+}0$  &	$+2$	&	$\phantom{+}0$  &  $\bar{m}^{f^{Y}}_{Y\backslash X}$		&	$-1$		&	$-1$		&	$+1$		&	$+1$   \\
			$\bar{m}^{f^{X\!Y}}_{X\!Y}$ 	&	$+1$ &	$-1$  &	$-1$ &	$+1$  &  $\bar{m}^{f^{X\!Y}}_{\backslash X\!Y}$ 	&	$-1$	&	$+1$ &	$+1$ &	$-1$ \\       
			\bottomrule
		\end{tabular}
	\end{center}
	\label{tab:binary_or}
\end{table}

The apparent paradox arises from the fact that the single feature relevances vanish if the other, conditioned variable is set to 1.
Explicitly, this means that $\bar{m}_{X|y}=0$  if $y=1$ as the outcome of $X \vee Y$ is already completely specified for $y=1$. 
The same applies to $\bar{m}_{Y|x}$ for $x=1$, from which \cite{strumbelj2010efficient,strumbelj2013explaining} incorrectly conclude that neither $X$ nor $Y$ are relevant for the prediction in this case. 
This apparent contradiction is obviously resolved by incorporating interaction effects, see \Cref{tab:binary_or}: Firstly, we note that all shielded main effects are positive (negative) for a value of one (zero). Secondly, the shielded joint effect is only positive in the exclusive or combination. In \Cref{app-sec:and_or_xor}, we additionally demonstrate that $X \vee Y$, $X \wedge Y$ and $X \veebar Y$ share the same shielded joint effects up to a constant factor, which has already been demonstrated on a global interaction level in \cite{lengerich2020purifying}. This at first sight slightly unintuitive result, illustrates the danger of inferring intuitive ground truth relevances and interactions for seemingly simple functions.

\subsection{Regression: Synthetic datasets} 	\label{subsec:syntheticdataset}

This section aims to validate the definitions for both single feature contribution and feature interaction based on a synthetic regression task. 
The main message we try to convey is that \pd{} successfully grasps the relevant contributions for a model-agnostic interpretation.

We consider a synthetic dataset with four independent features $\mathcal{X}=\{X^a, X^b,X^c,X^d\}$, generated by a Gaussian distribution with mean zero.
Additionally, we defined a target function:
\begin{equation}
	f(\mathcal{X}) = (X^a)^2 + 3X^b + \sin(\pi X^c) - \frac{(X^d)^3}{2} + 2\,\text{sgn}\,(X^a)\,\text{abs}(X^b).
	\label{eq:targetfunction}
\end{equation}
At this point, we want to stress that this choice is rather arbitrary. However, we believe that the results and conclusions are generic and invite the reader to try different functional forms in the accompanying notebook. 
In this section, we present results for a \textit{Random Forest} regressor trained on 3600 samples.

\begin{figure}[]
	\centering
	\begin{tabular}{c c}
		\small{(a) \pd{} relevances}	&  \small{(b) shielded main effects}	\\
		\includegraphics[width=.475\linewidth]{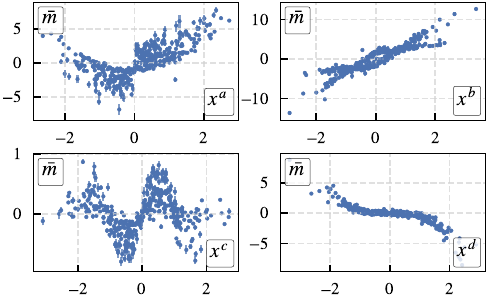} &	\includegraphics[width=.47\linewidth]{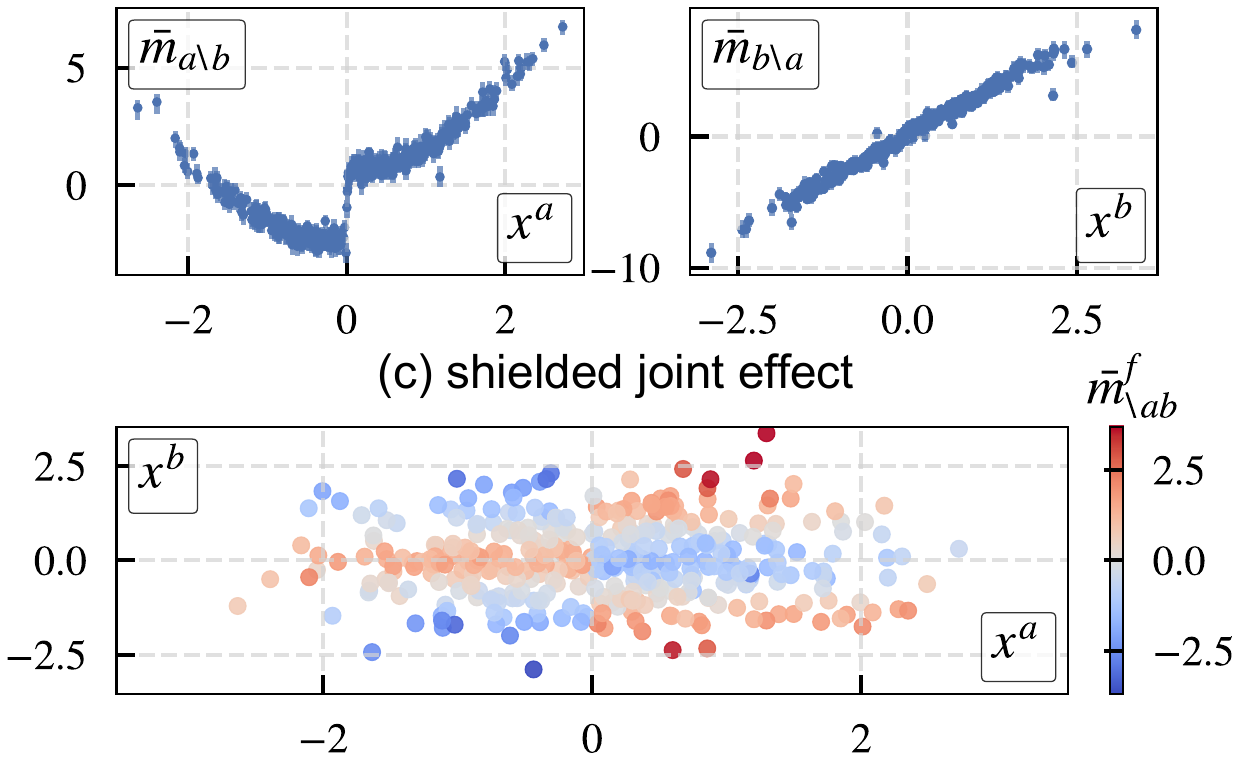} \\
	\end{tabular}
	\caption{\pd{} attributions for synthetic regression task (using the analytically known imputer distribution). 
		(a):~~\pd{} relevances as given by \Cref{eq:defmbar}. All additive terms are successfully captured, however, including the interaction between $X_a$ and $X_b$. 
		(b): shielded main effects,  \Cref{eq:completenessshielded}. The discontinuity at $x_a = 0$ is caused by a monotonic positive (or negative) interaction contribution. 
		(c): color-encoded shielded joint effect of feature $X_a$ and $X_b$. Interaction is given by $\text{sgn}(X_a)\,|X_b|$.
		\pd{} naturally provides uncertainty estimates for relevances via bootstrapping ($\#=200$ imputations).
		These results are in qualitative agreement with the Shapley Interaction Index \cite{lundberg2020local}, see \Cref{app-sec:syntheticdataset} for an explicit comparison.}
	\label{fig:synthetic_dataset}
\end{figure}

In \Cref{fig:synthetic_dataset} we show a possible workflow for the interaction analysis: On the left side, the raw \pd{} attributions are shown. 
We observe that all individual contributions are recovered correctly, e.g., the sinusoidal ($X^c$) or cubic ($X^d$) functional form is immediately recognizable.
Examining feature $X^a$ and $X^b$ next, we observe additional structure superimposed onto the underlying raw additive feature contribution. In particular, the main effect for $X^b$ shows two distinct branches, which clearly indicates the presence of an interaction.
Importantly, single feature attribution methods are restricted to this analysis depth. 
However, investigating the \pd{} joint effect of $X^a$ and $X^b$, allows us to go one step further. To this end, we show the shielded \pd{} attribution given by \Cref{eq:completenessshielded} on the right side of \Cref{fig:synthetic_dataset}.
In the top panel, the shielded main effects are shown. They correspond to the feature contribution without specifying the other feature. Consequently, only the pure additive feature contribution remains. 
In the lower panel, we show the color-encoded shielded joint effect of $X^a$ and $X^b$. Here, we immediately recognize the absolute value contribution of $X^b$ combined with a sharp transition at $x^a =0$ induced by the sign operation.
The latter also explains the spurious jump at the origin in the shielded main contribution of $X^a$. Due to the sign operation, feature $X^a$ has a monotonic positive or negative effect. Hence, a part of the interaction term can solely be contributed to feature $X^a$ and causes the discontinuity at the origin. 
Please note that this is in contrast to feature $X^b$, for which all interaction contributions are removed and only the linear dependence remains. These results, both for main as well as joint effects, are in qualitative agreement with the results obtained with the Shapley Interaction Index \cite{lundberg2020local}, see \Cref{app-sec:syntheticdataset}. 
Finally, in \Cref{app-sec:correlation}, we also repeat the experiment with correlated Gaussian features. This setting reveals differences between both approaches: Whereas Shapley distributes relevance evenly in the limiting case of perfectly correlated features, \pd{} single feature relevances tend to zero in this case. 
This can be seen as a sign for higher reliability of \pd{} relevances, since positive/negative attributions are guaranteed to be caused by the model and are not inflated by the conditional dependence.

In summary, \pd{} has successfully disentangled all relevant contributions. Importantly, and in contrast to, e.g., the Shapely interaction index, there was no need to calculate all possible interactions.
By manual inspection, we could select the relevant features and calculate the shielded effects with linear computational costs. 
From a more general point of view, this touches upon the problem of efficiently identifying interacting feature sets and potentially combining them in a hierarchical fashion, see \cite{Jin2020Towards,tsang2020does} for approaches in this direction, which we leave as future work.

\subsection{Regression: Real-world dataset NHANES}\label{sec:nhanes}

To demonstrate that \pd{} can also be applied to real-world regression datasets, we revisit the NHANES dataset \cite{cox1997plan}, which was discussed at length in \cite{lundberg2020local}. It is a healthcare dataset with 14,407 individuals. The prediction task is to infer the (log relative) risk of mortality based on 79 features. 
We train a \textit{Random Forest} and compute relevances for all individual features via \Cref{eq:defmbar}. 
Here, we use the (conditional) Mahalanobis imputer \cite{aas2021explaining} and show results for the (marginal) train set imputer in \Cref{app-sec:nhanes}.

The results on feature relevances are shown in the top panel of \Cref{fig:nhanes_relevances}. We infer global feature importances by computing the mean of the absolute relevance score for each feature across the whole test set. 
The three most important attributes agree with previous investigations based on the SHAP TreeExplainer \cite{lundberg2020local}, although the ordering of the features sex and systolic blood pressure is interchanged. 
Also, the single feature relevances for the two most important continuous features age and systolic blood pressure are in qualitative agreement with earlier investigations.

\begin{figure}[]
	\centering
	\begin{tabular}{ccc}
		& (a) Relevance & \\
		\includegraphics[width=0.33\textwidth,align=c]{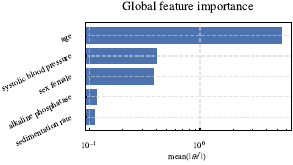} &
		\includegraphics[width=0.3\textwidth,align=c]{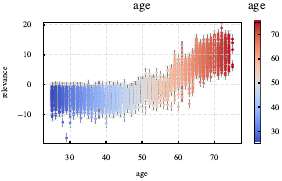}	&
		\includegraphics[width=0.3\textwidth,align=c]{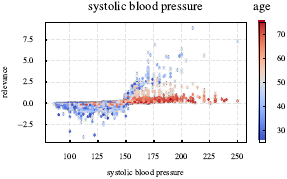}	\\ \noalign{\smallskip}
		& (b) Interactions & \\

		\includegraphics[width=0.33\textwidth,align=c]{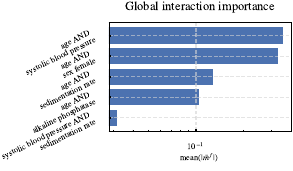} &
		\includegraphics[width=0.3\textwidth,align=c]{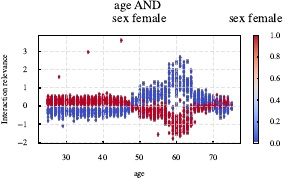}	&
		\includegraphics[width=0.3\textwidth,align=c]{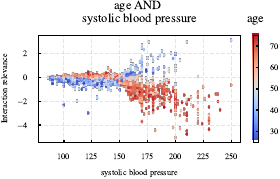}
	\end{tabular}
	\caption{\pd{} (interaction) relevances for a \textit{Random Forest} trained on the NHANES dataset using a (conditional) Mahalanobis imputer \cite{aas2021explaining}, that are in qualitative agreement with existing methods \cite{lundberg2020local}. 
		top left:~~Ranking of the five most important features. 
		top center:~~Relevance for the most important feature age.
		top right:~~Relevance for the second most important feature systolic blood pressure.
		bottom left:~~Ranking of the five most important feature interactions.
		bottom center:~~Interaction relevance between systolic blood pressure and age.
		bottom right:~~Interaction relevance between age and sex, revealing a pronounced age dependence.}
	\label{fig:nhanes_relevances}
\end{figure}

Next, we turn to interaction relevances in the bottom panel of \Cref{fig:nhanes_relevances}. Similarly, as for feature relevances, we assess the global interaction relevance from the mean absolute interaction relevances across the whole test set. 
Here, we consider all pairs of interactions between the five most important features identified in the first step.
Systolic blood pressure and age, as well as age and sex, show pronounced interaction effects and the corresponding interaction relevances on a per-sample basis are again in qualitative agreement with literature results \cite{lundberg2020local}.

\subsection{Classification: MNIST}	\label{subsec:mnist}
In the previous sections, we have established intuitive global interpretations using local \pd{} attributions. In this section, we move forward and analyze instance-wise attribution maps. 
To showcase the abilities of \pd{}, we use the MNIST dataset, which allows for an intuitive interpretation of the resulting attribution maps and is not too small neither in terms of dataset size nor in terms of input dimensionality. 
We train a fully-connected classifier (hidden layers 1000 and 500) and achieve an accuracy of $97.8\%$  after $n_\text{epochs} = 10$ epochs of training. 
To enforce a proper probabilistic interpretation, we calibrate the network using temperature scaling as proposed in \cite{Guo2017Calibration}. This is the natural way of dealing with potential saturation issues without the need to adjust the original formalism as in \cite{gu2020contextual}. 
\begin{figure}[]
	\centering
	\begin{tabular}{l || r}
		\includegraphics[width=.34\columnwidth]{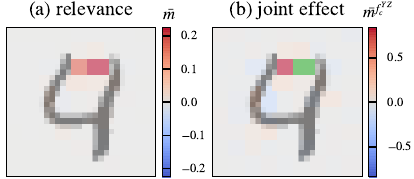}  &
		\includegraphics[width=.34\columnwidth]{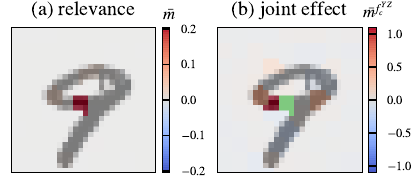}\\ 
		\includegraphics[width=.34\columnwidth]{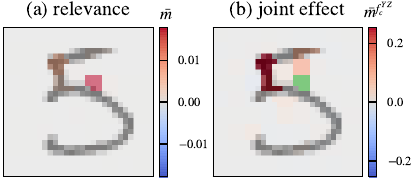} &
		\includegraphics[width=.34\columnwidth]{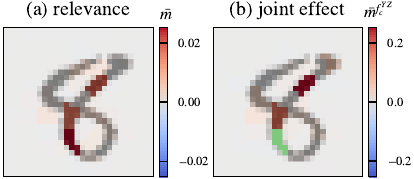} 	
	\end{tabular}
	\caption{\pd{} for MNIST digits calculated on $\sim50$ SLIC superpixels \cite{achanta2012slic} using a (conditional) vae imputer for the true (and correctly predicted) class label. (a)~~\pd{} relevances/main effects, \Cref{eq:defmhat} (b)~~\pd{} joint effects, \Cref{eq:rel_sum}, with respect to the marked (green) reference super-pixel of highest relevance. Further examples can be found in \Cref{app-sec:mnist}. We used $\#=600$ imputations.}
	\label{fig:mnist}
\end{figure}

\subsubsection{Meaningful relevance and interaction attributions}
We analyze this calibrated model using \pd{} in the next step. To this end, we obtain $\sim50$ superpixels via the Simple Linear Iterative Clustering (SLIC) algorithm \cite{achanta2012slic}.
However, \pd{} gives no restrictions on this selection, see \cite{Tian2017,Wei2018Explain} for similar approaches and \Cref{app-sec:mnist} for results with more finegrained superpixel}. Importantly, this flexibility is retained by our novel interaction measure. 
Here, we use a (conditional) variational autoencoder (VAE) imputer with pseudo-Gibbs sampling \cite{pmlr-v32-rezende14}, see \Cref{app-sec:imputer} for details. Additionally, we show results for a (marginal) train set imputer in \Cref{app-sec:mnist}.
In \Cref{fig:mnist}~(a) we show the corresponding attributions for four different digits. The digits are chosen to be a representative subset of the complete test set. We see that the attributions are visually reasonable, e.g., the characteristic white space for the four and five are highlighted or also the characteristic parts of figure eight and nine. This demonstrates \pd{}'s ability to produce intuitively meaningful attributions.

In the next step, we analyze the interaction measure. To this end, we calculate the joint effect between all super-pixels with respect to the super-pixel with highest relevance and show the resulting heatmap in \Cref{fig:mnist}~(b).
The first thing to note is that the heatmaps are sparse and hence, informative. Our measure clearly highlights the intuitively related figure parts such as neighboring pixels.
In contrast, if we would measure the overall effect of both pixels, the resulting heatmap would be blurry and covered up by the main effects. 
Additionally, we note that joint effects are particularly pronounced for meaningful combination of superpixels. For example, consider the digit five, here the \emph{enclosing corner} is highly connected to the characteristic (reference) white space. 
This means that the model jointly leverages the information of both superpixels, i.e., a corner combined with an open whitespace is likely a five. Similar conclusions can be drawn from the other digits, e.g., the digit four is characterized by the centered whitespace enclosed with a vertical stroke.

{\subsubsection{Comparison to Shapley Interaction Index}}

\begin{figure}[]
	\centering
		\noindent
	\begin{tabular}[t]{l| c | c}
	\multicolumn{1}{c}{}	&  (a) Relevances		&			(b) Interactions	\\
	\pd{} & 	\includegraphics[width=.4\columnwidth]{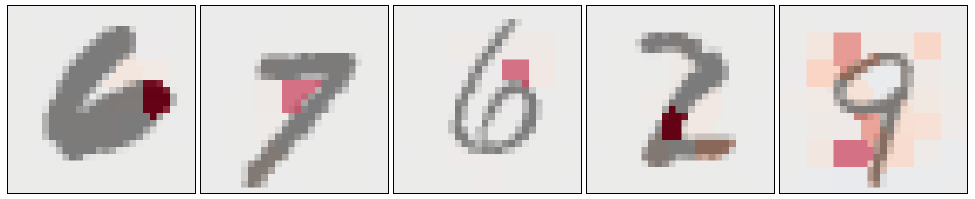}				&			\includegraphics[width=.4\columnwidth]{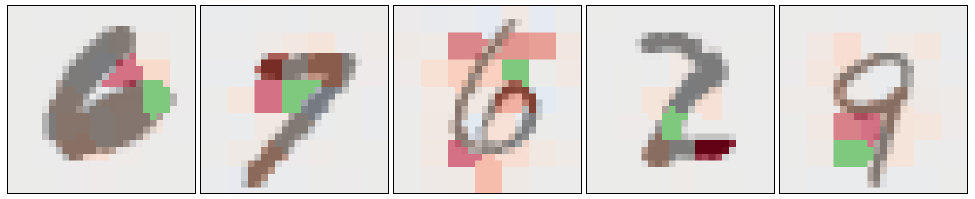}			\\ 
	Shapley	&		\includegraphics[width=.4\columnwidth]{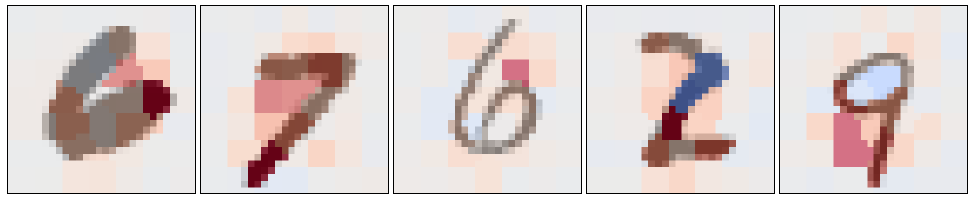}				&				\includegraphics[width=.4\columnwidth]{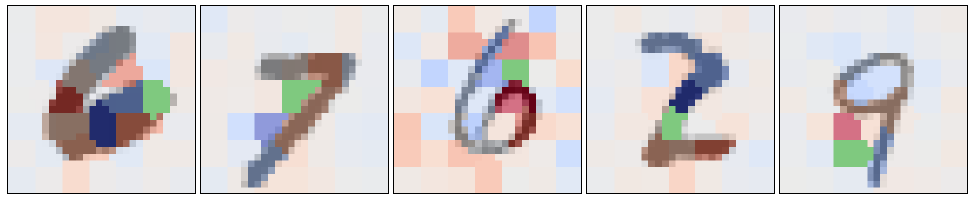}	\\
\end{tabular}
	\caption{Comparing \pd{}  relevances/joint effect and Shapley values/interaction index on randomly selected digits for the true (and correctly predicted) class label using a (conditional) VAE imputer. Interaction measured with respect to the marked (green) reference super-pixel of highest relevance. \pd{} and Shapley values produce qualitatively similar feature and interaction attributions based on $\# = 600$ model calls per attribution. }
	\label{fig:mnist_shapcomparison}
\end{figure}
We now compare the \pd{} joint effect to the Shapley Interaction Index \cite{lundberg2020local}, which directly builds upon the popular Shapley value concept. 
To this end, we use a custom Shapley implementation, which approximates the true Shapley values via subsampling coalitions $S$. This approach was proposed in \cite{strumbelj2010efficient} for traditional Shapley values and can straightforwardly be extended to calculate the Shapley Interaction Index. 
Importantly, our custom implementation allows comparing \pd{} vs. Shapley values based on identical (conditional or marginal) imputer distributions. 
To ensure a consistent comparison, we need to account for a global sign between the Shapley Interaction Index and the \pd{} raw joint effect, cf. \Cref{eq:shapleyinteractionindex} for further details.

Within this setting, we provide attributions for randomly selected examples in \Cref{fig:mnist_shapcomparison}. 
We find qualitative agreement between \pd{} and Shapley value attributions, both in terms of relevances and in terms of interaction measures.
It is worth stressing that both relevance attributions are generally aligned. Due to the close relationship between \pd{} and Shapley values, both approaches allow for similar qualitative insights.
	Interestingly, in most cases for which the relevances do not fully align, the difference between both heatmaps is at least partially compensated for by the corresponding \pd{} interaction effect. This observation aligns with the \emph{completeness axiom} discussion  in \Cref{sec:efficiency} and potentially allows for a low-cost Shapley approximation based on the \emph{interaction completeness} relation. Particularly, the latter needs to be investigated in a dedicated follow-up study.
These findings are robust against using a (marginal) train set imputer as shown in \Cref{app-sec:mnist}.
In summary, \pd{} joint effects are capable of extracting information on feature interactions in a scalable and model-agnostic fashion. 
Importantly, this kind of analysis can easily be extended to large-scale image datasets, see \cref{subsec:cub_birds}.

\subsubsection{\pd{}'s superior computational scaling} \label{sec:superior_num_scaling}
\begin{figure}
		\includegraphics[width=\linewidth]{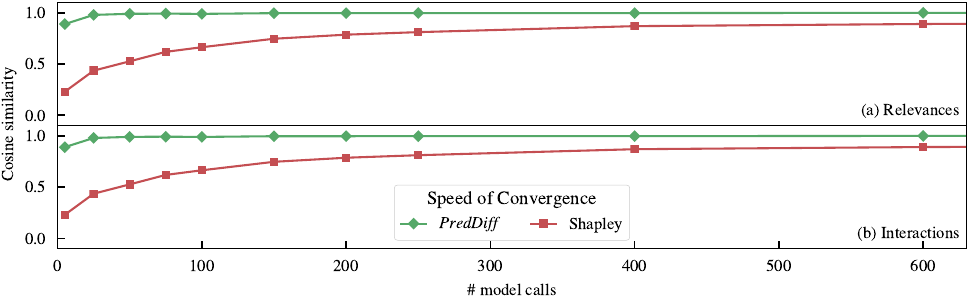}
	\caption{Analyzing the convergence speed of \pd{} attributions compared to Shapley values. Approximation fidelity measured via the cosine similarity between a high-fidelity baseline $\bar{m}^\infty$/$\phi^\infty$ ($\#=1200$) and an approximate heatmap $\bar{m}^{\#}$/$\phi^{\#}$, which used $\#$ model calls. Experiments are based on 50 random MNIST test samples and a (marginal) train set imputer. 
	Interaction measure with respect to the reference super-pixel of highest \pd{} relevance (equivalent to \Cref{fig:mnist_shapcomparison}). }
	\label{fig:computational_scaling}
\end{figure}

We now investigate the numerical fidelity of \pd{} and compare it to Shapley values. Previously in \Cref{sec:scaling}, we theoretically established that \pd{} provides the optimal linear scaling in terms of model calls \#. Next, \Cref{fig:computational_scaling} experimentally supports this claim. Therein, we compare the numerical convergence properties of \pd{} vs. Shapley values both for relevance and interaction attributions.
To this end, we first compute a numerically expensive, high-fidelity baseline $\bar{m}^\infty$/$\phi^\infty$ for both approaches using $\# = 1200$ model calls. For this reason, we restrict ourselves to the (marginal) train set imputer in this particular experiment. However, as the previous comparison around \Cref{fig:mnist_shapcomparison} indicates, these findings straightforwardly generalize to other imputers. For interactions, we stick to the comparison established around \Cref{fig:mnist_shapcomparison} and calculate an interaction heatmap with respect to the super-pixel of highest \pd{} baseline relevance. 
Subsequently, these high-fidelity baselines $\bar{m}^\infty$/$\phi^\infty$ are compared to approximate heatmaps $\bar{m}^{\#}$/$\phi^{\#}$, which are based on \# model calls. 
We measure the approximation fidelity via the cosine similarity with respect to the flattened heatmaps. Consequently, a cosine similarity of one reflects optimal alignment, i.e., perfect convergence, whereas lower values indicate  noisy attributions. 
From \Cref{fig:computational_scaling} it is clear that \pd{} attributions converge rapidly to the high-fidelity baseline ($\#\leq50$). In contrast, Shapley values do not fully converge and are limited to a noisy baseline approximation. Importantly, these findings are independent of whether one considers relevance or interaction attributions. Arguably, this is fundamentally related to the necessity of sampling all possible coalitions $S$, which is a possible source of numerical noise. In summary, \pd{} attributions are less noisy and effectively easier to access in real-world applications.

The previous results indicate that \pd{} rapidly converges towards its own high-fidelity baseline, but do not allow for conclusions on the quality of the resulting attributions. Inspired by \cite{crabbe2021explaining, ismail2020benchmarking}, we report in \Cref{sec:dynamask} on a synthetic experiment where the ground truth relevances are known by construction. Here, \pd{} main and joint effects show a considerably better overlap with the ground truth attributions as compared to sampled Shapley values, irrespective of the number of model evaluations.

\subsection{Classification: CUB Birds} \label{subsec:cub_birds}
\begin{figure}
	\centering
	(a) High relevance reference superpixel 
	\includegraphics[width=\linewidth]{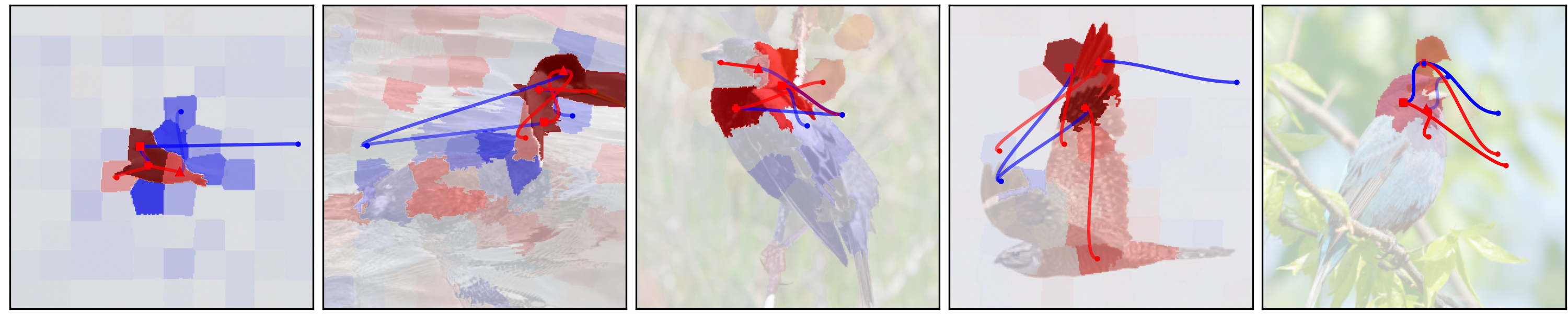}
	(b) Random reference superpixel
	\includegraphics[width=\linewidth]{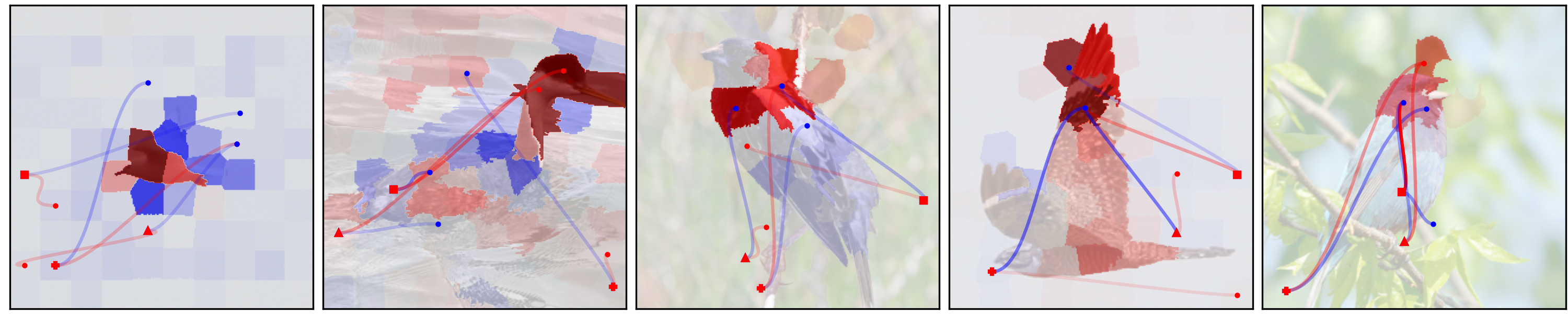}
	\caption{Interaction for the CUB birds dataset for 5 randomly selected samples obtained using a (conditional) histogram imputer \cite{Wei2018Explain}. The transparency of the red~(blue) b\'ezier curves represents the strength of the most positive (negative) raw joint effects. The \pd{} relevances are visualized as heatmaps.
	In the upper panel the three reference superpixel are chosen according to highest relevance. In contrast, the lower panel shows random reference superpixels. The predicted class probabilities of the five samples are 0.15, 0.62, 0.89, 0.856 and 0.98.
	}
	\label{fig:cub}
\end{figure}
As a proof-of-concept to demonstrate that \pd{} is applicable to high-resolution, real-world datasets, we present results on the CUB-200-2011 birds dataset \cite{WahCUB_200_2011}. More specifically, we finetune a vgg16 \cite{Simonyan15} model that was pretrained on ImageNet on the CUB dataset while excluding a small number of overlapping samples from the CUB test set. As for \pd{}, we work with superpixels determined using the Simple Linear Iterative Clustering (SLIC) algorithm \cite{achanta2012slic} and with a (conditional) histogram imputer \cite{Wei2018Explain}. A dedicated study on the imputer dependence of the results is deferred to future work.

In \Cref{fig:cub}, we show the results for five randomly selected test set samples. In the top row, we visualize the two most positive (negative) raw joint effects for the three reference superpixels with highest relevance.
This is to be contrasted with the bottom row, where three random reference superpixels are chosen. First-of-all, the results reveal that interaction effects do exist. These cannot be captured by the predominantly used single-pixel attribution methods, which implicitly distribute them onto single-feature relevances \cite{deng2020unified}.
As an interesting observation, the largest interactions occur between the individually most relevant superpixels.
On the contrary, the interaction between random superpixels typically remains small. As one expects, these random superpixels do not show a joint effect on the model prediction.
Strong interactions between spatially separated superpixels, which are visible in several examples in the top row, could be interpreted as signs for more complex reasoning patterns, which remain to be uncovered in detail in the future.
We close by stressing that the direct measurement of interaction effects in large-scale datasets, such as the CUB dataset, is impossible with most competing attribution methods, which show a less favorable scaling compared to \pd{}.
\section{Conclusion}
In this work, we revisited \pd{} as a model-agnostic attribution method that is firmly rooted in probability theory. We carefully analyze its theoretical properties and demonstrate its close relation to Shapley values.
Both rely on the same foundations but \pd{} only evaluates a minimal subset of terms considered by Shapley values. This enables a favorable linear scaling behavior. 
The main focus of our investigation lies in the analysis of feature interactions.
Here, we present an \emph{interaction completeness} property, which allows decomposing the relevance--for a given set of features--into main effects and joint (interaction) effects.
Crucially, this enables a targeted in-depth analysis to substantially increase model understanding.
Secondly, we shed new light on the foundations of model-agnostic interpretability methods for classification and propose a novel argument based on the \emph{no-interaction} property. In conclusion, the argumentation clearly favors logarithmic differences as the appropriate attribution measure, as it correctly disentangles the conditional classifier distribution from the underlying data distribution.
We discuss consequences for both \pd{} and Shapley values. For the reader's convenience, we concisely summarize the main properties and advantages of \pd{} in \Cref{app-sec:summary}.

In our experiments, we demonstrate how interaction effects can resolve apparent paradoxes and lead to a better understanding of the model behavior. 
Due to the favorable scaling of \pd{}, for both relevances as well as interaction measures, it is applicable in real-world scenarios.
As a first step in this direction, we analyze the interaction effects for an image classifier. The results clearly indicate that the classifier jointly exploits different image patches.
These in-depth insights are not possible via conventional feature-wise attribution methods. 
The foundations laid out in this work, pave the way towards  systematic investigations of interaction effects in more realistic use-cases and datasets. 
From our point of view, a sensible next step in this direction would be a systematic study of the imputer dependence on both relevances and \pd{} joint interaction effects on a large image dataset such as ImageNet.

\appendix

\section{Summary: Main properties of \pd{}}
\label{app-sec:summary}
A code repository to reproduce the experiments reported in the main text can be found at \url{https://github.com/AI4HealthUOL/preddiff-interactions} \\
We summarize the most important properties of \pd{}: 
\begin{itemize}
	\item \textbf{Conceptual simplicity:} 
	For \textit{well-calibrated} classifiers, \pd{} is deeply \textit{grounded in probability theory}, see \Cref{eq:defmclas}. 
	Additionally, interaction effects provide a novel argument in favor of \emph{logarithmic differences}, as relevance measure.
	
	\item \textbf{Arbitrary feature sets:} \pd{} can adaptively evaluate relevances for \textit{arbitrary sets of features}. These relevances naturally \textit{include all interaction effects} (i.e., are inherently non-additive).
	
	\item \textbf{Error estimates:} \pd{} provides an \textit{uncertainty estimate for relevances} on a per-sample basis via bootstrapping.
	
	\item \textbf{Imputation/On-manifold:} The imputation process, which is a necessary component of all perturbation-based approaches, is completely transparent through an \textit{exchangeable imputer}. In addition, using conditional rather than marginal probabilities for imputation alleviates the common problem of evaluating the classifier far from the data manifold.
	
	\item \textbf{Linear Scaling:} Most crucially for practical applications, both  \pd{} relevances and interactions enjoy a \textit{linear scaling with the number of feature sets} for which relevances/interactions are supposed to be evaluated.	
	The scaling coefficient can readily be adjusted by varying the number of imputations, see \Cref{app-fig:dependence_N_imputation}.
	Additionally, in practical applications often semantically meaningful feature combinations, rather than individual single features themselves, are the true objects of interests \cite{Wei2018Explain}.
	
	\item \textbf{Quantifying interaction effects:} \pd{} provides a \emph{decomposition formula} for relevances into main and joint effects, see \Cref{eq:rel_sum} and \Cref{app-eq:completeness_threepoint} for the generalization beyond two feature sets, in the form of an \emph{interaction completeness} property.
\end{itemize}

\section{Approximation using finite samples}
\label{app-sec:sampleapproximation}
\begin{figure}
	\begin{tabular}{c c}
		$N=5$		&	$N=50$\\
		\includegraphics[width=.47\linewidth]{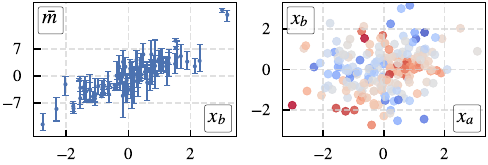}	&
		\includegraphics[width=.47\linewidth]{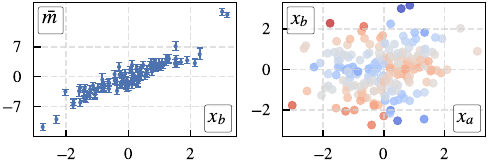}	\\			
	\end{tabular}
	\caption{Visualizing computational dependence of \pd{}; $N$: number of imputations. 
		LEFT:~~\pd{} relevances $\bar{m}^f_{x_b}$ according to \Cref{eq:defmbar} and using \Cref{app-eq:unified_integral}. High statistical accuracy needed to resolve branching caused by the interaction. 
		RIGHT:~~color-encoded shielded joint effect $\bar{m}^{f_{A\!B}}_{\backslash A\!B}$. The color bar is approximately equal to \Cref{fig:synthetic_dataset}.}
	\label{app-fig:dependence_N_imputation}
\end{figure}

The \pd{} relevances, \Cref{eq:defmclas}, can be approximated by sampling from the respective conditional distributions, i.e.,
\begin{equation} \label{app-eq:unified_integral}
	\begin{split}
		m^f_{Y|x}&=\int f(x,Y) p(Y|x) \text{d} Y\approx \sum^N_{j:y_j \sim p(Y|x)} f(x,y_j)\,,
	\end{split}
\end{equation}
for a potentially multidimensional $Y$. As discussed in the main text, there are many perturbation-based attribution methods that can be understood as single sample ($N=1$) approximations of \pd{}. 
In \Cref{app-fig:dependence_N_imputation} we show that a general trend is easily recovered with few samples, but more samples are needed for high fidelity attributions. Importantly, suppressed interaction signals are immediately visible via measuring the joint effect of features. 
In contrast to other attribution methods, \pd{} offers meaningful error bars without any additional overhead via bootstrapping. This is particular important to balance the trade-off between statistical accuracy and computational costs.  

Turning to the interaction relevance,  \Cref{eq:rel_sum}. Here, we first consider a regression setting, for which we can rewrite the joint effect in a numerically more convenient form, i.e.,
\begin{align}
	\begin{split}
		\bar{m}^{f^{Y\!Z}}_{Y\!Z|x} &= m^{f^Y}_{Y|x} + m^{f^Z}_{Z|x} - m^{f}_{Y\!Z|x} + f(x, y, z) \\
		&= \int \text{d}Y\text{d}Z \Big\{\big[f(x, Y, z) + f(x, y, Z) - f(x, Y, Z)\big]  p(Y, Z|x) \Big\} - f(x, y, z)		 \,.	
	\end{split}
\end{align}
This identity allows reusing imputations for every $m$-value evaluation and consequently, reduces numerical noise significantly. 

We now turn to a classification setting. Here, we are bound to explicitly intervene on feature $Y$ and $Z$ and break their dependence. In \Cref{sec:classification} we propose to sample from the joint distribution, e.g., $Y, Z\sim q(Y,Z|x)$, for all centered $m$-values in \Cref{eq:rel_sum}. For the main effects, one discards the redundant features $Y$ or $Z$. For the joint effect, one intervenes and shuffles $Y$ and $Z$. Thereby, one samples from the distribution $q(Y|x)q(Z|x)$.

\section{Imputation algorithms}
\label{app-sec:imputer}
In this work, we make use of the following imputation algorithms:
\paragraph{Train Set Imputer:} The \textit{Train Set Imputer} uses randomly sampled instances from the training set to impute respective values in the target features.
This was among the imputers proposed in the original \pd{} publication \cite{Sikonja2008Explaining}.
Along the line our discussion in \Cref{sec:shapleyinteraction}, we employ an factorizing train set imputer distribution, i.e., each segment is imputed with an independent train set sample.
\paragraph{Mahalanobis Imputer:} The \textit{Mahalanobis Imputer} \cite{aas2021explaining} can be seen as a generalization of the \textit{Train Set Imputer}. It also returns training set samples of the respective features to be imputed but additionally provides a weighting factor. These weights are obtained from a kernel estimator based on the Mahalanobis distance. 
\paragraph{Multivariate Gaussian Imputer:} The \textit{Multivariate Gaussian Imputer} samples from a multivariate, conditional Gaussian distribution that is conditioned on the values of the features that are not to be imputed. In a \pd{} application in computer vision, a similar imputer was used in \cite{zintgraf2017visualizing}.
\paragraph{Variational Autoencoder with Pseudo-Gibbs Sampling:} A trained variational autoencoder can be used for imputation by iteratively passing the sample through encoder and decoder.
After each iteration values of features not to be imputed are restored. This procedure was shown to approximately sample from the desired conditional distribution \cite{pmlr-v32-rezende14}. In the MNIST example, we use fully connected encoders and decoders each with hidden units 500 and 256.
\paragraph{Color Histogram Imputer:} The \textit{Color Histogram Imputer} was introduced in \cite{Wei2018Explain} and is based on sampling from the colors present in the image. To this end, one generates a histogram of all RGB values within an image and subsequently, imputes with a color sampled from this histogram, which is interpreted as a probability distribution. Importantly, the imputed patches are uni-color.

\section{Properties of \pd{} relevances and interactions}
\label{app-sec:preddiff_properties}

\subsection{Properties of \pd{} relevances:}
We discuss basic properties of \pd{} relevances based on the five axioms investigated in \cite{sundararajan2017axiomatic}. In particular, these include the classic Shapley axioms \cite{shapley1953value} \emph{completeness}, \textit{linearity}, \textit{symmetry} and \textit{null player}. Properties of attribution methods are typically investigated in a regression setting and not investigated in a classification setting. 
The \pd{} formalism provides an explicit definition of the relevance in terms of calibrated class-wise output probabilities and therefore, allows verifying properties explicitly in the classification setting.

\subsubsection*{Completeness/Efficiency/Additivity/Local accuracy}\noindent
The \textit{completeness axiom} states that the summed relevances $\phi_{i}^f$ of all individual features $i$ should yield the difference between the function value and a reference value, $f(x) = \phi_0^f + \sum_i \phi_i^f$. In the \pd{} framework, relevances for individual features are not distinguished compared to those of arbitrary combinations of features. In particular, there is no reference value, which is either set explicitly as for Integrated Gradients \cite{sundararajan2017axiomatic} or implicitly as for Shapley-values. In contrast, for every sample and feature combination there is a separate reference point for which the relevance vanishes. Note that the \emph{completeness axiom} is satisfied for linear models with independent features, see \Cref{app-sec:linear}. Finally, it is worth stressing that the \emph{completeness axiom} is recovered from \pd{}'s \emph{interaction completeness} property upon including all interaction effects, see \Cref{sec:efficiency} for a detailed discussion.

\subsubsection*{Sensitivity/Dummy/Null Player/Missingness} \noindent
 Consider a function $f(X,Y)=f(X)$ that does not depend on the features $Y$. We find
\begin{equation}
	m_{Y|x}^f = \int f(x) p(Y|x) \text{d}Y = f(x),
\end{equation}
and hence, $\bar{m}_{Y|x}^f=0$, i.e., if $f$ does not depend on $Y$ also the corresponding relevance is zero. This property holds both for classification and regression.

\subsubsection*{Linearity} \noindent
 For regression, one easily verifies that $\bar{m}^{a f_1+b f_2}_{Y|x}=a \cdot\bar{m}^{f_1}_{Y|x}+b\cdot\bar{m}^{f_2}_{Y|x}$. In a classification setting, linearity in the output probabilities themselves is not a natural assumption and the property is also not satisfied. Note, that even for factorizing functions, i.e., additive log probabilities, the relevances in general do not decompose into two separate contributions.

\subsubsection*{Symmetry} \noindent For a function $f(X,Y,Z)$ that is symmetric with respect to exchanging $Y$ and $Z$, one easily verifies that also the relevances coincide, i.e., $\bar{m}^f_{Y|x,z}=\bar{m}^f_{Z|x,y}$ if evaluated at $y=z$and provided that also the data distribution $p(X,Y,Z)$ shares the same symmetry with respect to exchanging $Y$ and $Z$. 
The additional requirement on the data distribution is unavoidable for approaches that explicitly depend on the data distribution, as also realized in \cite{pmlr-v108-janzing20a,sundararajan2020shapley} in slightly different contexts. This property holds both for classification and regression.

\subsubsection*{Implementation Invariance} \noindent
The relevance is trivially independent of the way the function $f$ is implemented, as \pd{} is model-agnostic and only depends on the model outputs.

\subsection{Properties of \pd{} interaction relevances/joint effects}
In this subsection, we discuss basic properties of the \pd{} interaction relevance. 

\subsubsection*{No Interaction}\noindent
In an additive regression setting, i.e., if $f$ decomposes into a sum of two terms $f(X,Y,Z)=g(X,Y)+h(X,Z)$, the joint effect between variables $Y$ and $Z$ vanishes $\bar{m}^{f^{Y\!Z}}_{YZ|x}=0$.

In a classification setting, we require a vanishing joint effect in the case of generalized \textit{informative conditional interactions}, as specified in \Cref{eq:defnointeraction}, where we additionally require a factorizing imputer distribution, i.e., $q(Y,Z|x)=q(Y|x)\cdot q(Z|x)$, see the discussion in \Cref{sec:classification}. In this case, one can show $\bar{m}^{f^{YZ}_c}_{YZ|x}=0$, see \Cref{app-sec:derivationinteraction} for a detailed derivation.

\subsubsection*{Null Player} \noindent 
If $f$ does not depend on $z$, by the null player property for \pd{} relevances, we find $\bar{m}^{f^Z}_{Z|x}=0$ and additionally $m^f_{YZ|x}=m^{f^Y}_{Y|X}$. Hence, we find also $\bar{m}^{f^{YZ}}_{YZ|x}=0$. This property holds both for classification and regression.

\subsubsection*{Linearity} \noindent 
In a regression setting, one easily verifies that $\bar{m}^{(a f_1+b f_2)^{YZ}}_{YZ|x}= a\cdot\bar{m}^{f_1^{YZ}}_{YZ|x} + b\cdot\bar{m}^{f_2^{YZ}}_{YZ|x}$. As in the case of the \textit{linearity} property for \pd{} relevances, linearity is not a sensible assumption in the classification and also not satisfied in the \pd{} formalism. 

\subsubsection*{Symmetry} \noindent 
By construction, the interaction relevance is symmetric with respect to its arguments, i.e., $\bar{m}^{f^{YZ}}_{YZ|x}=\bar{m}^{f^{ZY}}_{ZY|x}$. This property holds both for classification and regression.

\section{\pd{} for linear models and elementary multiplicative interactions} 
\label{app-sec:linear}
\noindent
It is insightful to compute \pd{} relevances for linear models, i.e.,
\begin{equation}
	f(X^1,\ldots,X^d) = \beta_0 + \beta_1 X^1 + \ldots + \beta_N X^N\,.
\end{equation}
For a given subset $S$ of features, one now straightforwardly evaluates $m$-values,
\begin{equation}
	m^f_{X_S|x_{\bar{S}}} = \sum_{j\in S} \beta_j \mathbb{E}_{p\left(X^j|x_{\bar{S}} \right) }[X^j] + \sum_{j\in \bar{S}} \beta_j x^j + \beta_0\,,
\end{equation}
where $\bar{S}$ is the complement set of features evaluated at the sample point $x_{\bar{S}}$. This leads to centered $m$-values/relevances of the form
\begin{equation}
	\bar{m}^f_{X_S|x_{\bar{S}}} \,\, =  \,\,\sum_{j\in S} \beta_j \left(x^j -  \mathbb{E}_{p\left(X^j|x_{\bar{S}} \right) } [X^j] \right)\,.
\end{equation}
For a single variable, i.e., $S=\{X^j\}$, this yields the relevance 
\begin{equation}
	\label{app-eq:linearsinglevar}
	\bar{m}^f_{X^j|x_{\bar{S} }} \,\, = \,\, \beta_j \left(x^j -  \mathbb{E}_{p\left(X^j|x_{\bar{S}} \right)}[X^j] \right)\,,
\end{equation} 
which is in line with the expectation that for linear models the relevance should scale with the corresponding coefficient of the variable under consideration (after appropriate centering).

In particular, this implies that for linear models \pd{} also satisfies the \textit{completeness axiom}, which is also the situation where it is most desirable. This also follows explicitly from \Cref{app-eq:linearsinglevar},
\begin{equation}
	\sum_j\bar{m}^f_{X^j | x_{\bar{\{j\}}}} \,\, = \,\, f(x^1,\ldots,x^N) - \bar{f}\,,
\end{equation} 
where we assumed independent features in order to obtain a constant reference value $\bar{f}=\beta_0 + \sum_{j=1}^N \beta_j \mathbb{E}_{p(X^j)}[X^j]$. It is worth noting that this very expression is also obtained within the formalism of Shapley values \cite{kumar2020problems}, which directly follows from the fact that Shapley values are uniquely characterized by satisfying sensitivity, linearity, symmetry and completeness.

We consider also the second explicit example from \cite{kumar2020problems}. Here, we consider the simplest multiplicative interaction,
\begin{equation}
	g(X^1,\ldots,X^N) = \prod_{i=1}^N X^i\,,
\end{equation}
again under the assumption of independent features as above. For a given subset $S$ of features, one now straightforwardly evaluates $m$-values,
\begin{equation}
	m^g_{X_S|x_{\bar{S}}}=\prod_{j\in S} \mathbb{E}[X^j]  \,\,  \prod_{j\in \bar{S}} x^j \,,
\end{equation}
and hence,
\begin{equation}
	\bar{m}^g_{X_S | x_{\bar{S}}}=\prod_{j\in S} \left(x^j- \mathbb{E}[X^j] \right) \,\, \prod_{j\in \bar{S}} x^j \,.
\end{equation}
As before, for a single variable, i.e., $S=\{X^j\}$, this yields the relevance $\left(x^j - \mathbb{E}[X^j]\right)\prod_{k\neq j} x^k$. In particular, for centered variables, we have $\prod_{k} x^k$, which, again, coincides with the result from the Shapley formalism \cite{kumar2020problems} up to a global factor. However, contrary to the argumentation in \cite{kumar2020problems}, we do not see it as a contradiction that all features $X^j$ obtain the same relevance as opposed to assigning a larger relevance to features with a larger absolute numerical value, as we are dealing with an inherent interaction effect that cannot be distributed in a simple fashion.

\section{Shapley values}
\label{app-sec:shapley}
\subsection{Classification}
We first give the Shapley values based on the regression value function \Cref{eq:defvaluefct_regression}
\begin{align}
	\begin{split}
		\phi_a(v^\text{reg}) &= \frac{1}{2} \left[p(c|x^a) - p(c) + p(c|x^a, x^b) - p(c|x^b)\right] \\ 
		\phi_b(v^\text{reg}) &= \frac{1}{2} \left[p(c|x^b) - p(c) + p(c|x^a, x^b) - p(c|x^a)\right].
	\end{split}
\end{align}
Since all terms appear additive, it is not clear how one should leverage the multiplicative \emph{no-interaction} property. 
Consequently, all single feature contributions remain mixed, which clearly highlights the need for a special treatment of classification tasks. \\

\subsection{Shapley interaction index}
\label{app-sec:shap_interaction}
We now move forward and consider how interactions are explicitly treated in the Shapley formalism. In \cite{lundberg2020local} the 'Shapley Interaction Index' is proposed, an interaction measure based on a game theory \cite{Fujimoto2006}. It is given by
\begin{equation}
	\phi_{i, j}=\sum_{S \subseteq \backslash\{i, j\}} \frac{|S| !(N-|S|-2) !}{2(N-1) !} \delta_{i j}(S)
\end{equation}
for $i\neq j$ 
\begin{equation}
	\delta_{i j}(S)=v(S \cup\{i, j\}) - v(S \cup\{i\}) - v(S \cup\{j\}) + v(S).
\end{equation}
In this section we restrict ourselves to simply evaluate this interaction measure with respect to the \emph{no-interaction} properties introduced in \Cref{sec:interaction}.

\subsubsection*{Regression}
Here, we consider the additive function $f(X,Y,Z) = h(X,Y) + g(X,Z)$ for which $Y$ and $Z$ are clearly non-interacting. In contrast to \pd{}, we need to restrict $X$ to a single feature for an analytically tractable analysis. 
We then have two possible subsets $S \in \{\emptyset, \{X\}\}$, for which we can calculate the interaction contribution 
\begin{equation}
	\begin{split}
		\delta_{Y\!Z}(\emptyset) = \int \text{d}X&\Big\{ h(X, y)\big[p(X|y, z)- p(X|y)\big]  + g(X, z)\big[p(X|y, z)- p(X|z)\big] \\
		& - \int \text{d}Y h(X, Y) \big[p(X, Y|z) - p(X, Y)\big]  - \int \text{d}Z g(X, Z) \big[p(X, Z|y) - p(X, Z)\big]\Big\}
	\end{split}
\end{equation}
and 
\begin{equation}
	\begin{split}
		\delta_{Y\!Z}(x) = & \int \text{d}Y h(x, Y) \big[p(Y|x) - p(Y|x, z)\big] + \int \text{d}Z g(x, Z) \big[p(Z| x) - p(Z|x, y)\big].
	\end{split}
\end{equation}
We observe that using different imputer distributions has a non-trivial effect on the resulting attribution. It is clear that using a interventional (marginal) definition for the value function would resolve this problem and lead to a vanishing interaction contributions.

\subsubsection*{Classification}
We consider a classifier $p(c|x, y, z)$ that obeys the no-interaction property \Cref{eq:defnointeraction}, i.e., $p(y, z| x,c)= p(y|x,c) p(z|x,c)$. The Shapley values are based on the classification value function \Cref{eq:defvaluefct_classification}. 
Otherwise we use the same setting as for regression and the interaction contributions yield
\begin{equation} \label{app-eq:shap_clf_int0}
	\begin{split}
		\delta_{Y\!Z}(\emptyset) &= \text{log}_2\left(p(c|yz)\right) - \text{log}_2\left(p(c|y)\right) - \text{log}_2\left(p(c|z)\right)+ \text{log}_2\left(p(c)\right) \\
		&= \text{log}_2\left(\frac{p(y,z|c)}{p(y|c) p(z|c)}\right)	+ \text{log}_2\left(\frac{p(y)p(z)}{p(y, z)}\right)
	\end{split}
\end{equation}
and 
\begin{equation}\label{app-eq:shap_clf_int1}
	\begin{split}
		\delta_{Y\!Z}(x)&= \bar{m}^{f_c}_{Y|xy} + \bar{m}^{f_c}_{Z|xy} - \bar{m}^{f_c}_{YZ|x} \\ 
		&=\log_2\left(\frac{p(y,z|x, c)}{p(z|x, c)p(y|x, c)}\right)+\log_2\left(\frac{p(z|x)p(y|x)}{p(y,z|x)}\right).\\
	\end{split}
\end{equation}
The contribution $\delta_{Y\!Z}(x)$ is identical to the joint \pd{} effect up to a global sign and a slightly different conditioning. We start by discussing \Cref{app-eq:shap_clf_int1}, where the first term vanishes due to the no-interaction property. 
The second term relates to the \emph{mutual information dilemma}, which we discuss in \Cref{app-sec:derivationinteraction}. 
However, unlike for \pd{}, for which only the analogue of \Cref{app-eq:shap_clf_int1} applies, the 'Shapley Interaction Index' produces a conditional independence condition with respect to all subsets $S$. In the given case, this means that $\delta_{Y\!Z}(\emptyset)$ introduces two additional conditions $p(y,z|c)=p(y|c)p(z|c)$ and $p(y,z|x)=p(y|x)p(z|x)$. The second term could in principle be avoided upon using a fully factorizing, marginal imputer distribution, which potentially leads to off-manifold evaluations.
However, $p(y,z|c)=p(y|c)p(z|c)$ remains as an additional constraint that has to be imposed for a non-interacting classifier in the Shapley case. In general, this condition is not fulfilled, thus, the 'Shapley Interaction Index' does satisfy the \emph{no-interaction} property in its most general form.

\section{Anchored decomposition and interactions}
\label{app-sec:projections}

\subsection{Two-point interactions} \label{app-sec:two_variable_decomposition}
In this section, we focus our discussion on the simplest non-trivial case, where we are interested in the quantification of interaction effects between two sets of features $Y=\{Y^1,\ldots,Y^l\}$ and $Z=\{Z^1,\ldots,Z^m\}$ in presence of the remaining features $X=\{X^1,\ldots,X^k\}$. We aim to decompose the model function into terms that depend only on subsets of the set of feature sets $\{X,Y,Z\}$. The anchored expansion\footnote{An alternative, related approach would be to use a functional ANOVA decomposition using $p(x,y,z)$ as weight to off-manifold evaluation, which would in principle provide a similar decomposition. However, the projection would require numerous high-dimensional integrations instead of function evaluations as in the case of the anchored decomposition with additional complications in the case of correlated features \cite{hooker2007generalized}. Both issues prevent the approach from being widely applicable in real-world applications.} from \cite{kuo2010decompositions} with anchor point $c=(x^1,\ldots,x^k, y^1,\ldots, y^l,z^1,\ldots,z^m)$ gives us a decomposition of the form 
\begin{equation}
	\label{app-eq:anchored1}
	f(X,Y,Z) = \sum_{V\subseteq X \cup Y \cup Z} f^V\,.
\end{equation}
Its terms are given by
\begin{equation} \label{app-eq:decomposition_terms}
	f^V(V) = \sum_{W \subseteq V } (-1)^{|V|-|W|} P_{\mathcal X \backslash W} f(X,Y,Z)\,, 
\end{equation}
where $f^V$ is only a function of features contained in the set $V$ and $P_V$ is the projection that freezes the features in $V$ at their anchor point values, e.g., $P_{Y}f(X,Y,Z)=f(X,y,Z)$.
It is the unique decomposition of this form that satisfies the annihilating property $P_{X^j}f^V=0$ for all $X^j\in V$.  It can be shown that the decomposition is minimal, meaning that it never introduces unnecessary terms \cite{kuo2010decompositions}. 

We can recombine the terms in \Cref{app-eq:anchored1} as follows 
\begin{equation}
	\label{app-eq:anchoredf}
	\begin{split}
		f(X,Y,Z) &= f^\varnothing + f^{X}(X) + f^{Y}(Y) + f^{Z}(Z) \\ &+ f^{X\!Y}(X,Y) + f^{X\!Z}(X,Z) + f^{Y\!Z}(Y,Z) + f^{X\!Z}(X,Z) + f^{X\!Y\!Z}(X,Y,Z)\, ,
	\end{split}
\end{equation}
where
\begin{equation}
	\begin{split}
		f^{X\!Y\!Z}(X,Y,Z) &= \sum_{V_X\cup V_Y\cup V_Z | V_X\subseteq X \land V_Y\subseteq Y\land V_Z\subseteq Z \land V_X \neq \varnothing \land V_Y \neq \varnothing \land V_Z \neq \varnothing } f^V\,,\\
		f^{X\!Y} &= \sum_{V_X\cup V_Y | V_X\subseteq X \land V_Y\subseteq Y \land V_X \neq \varnothing \land V_Y \neq \varnothing } f^V\,,\\
		f^{Y\!Z} &= \sum_{V_Y\cup V_Z | V_Y\subseteq Y \land V_Z\subseteq Z \land V_Y \neq \varnothing \land V_Z \neq \varnothing } f^V\,,\\
		f^{X\!Z} &= \sum_{V_X\cup V_Z | V_X\subseteq X \land V_Z\subseteq Z \land V_X \neq \varnothing \land V_Z \neq \varnothing} f^V\,,\\
		f^{X} &= \sum_{V\subseteq X \land V \neq \varnothing } f^V\,,\\
		f^{Y} &= \sum_{V\subseteq Y \land V \neq \varnothing } f^V\,,\\
		f^{Z} &= \sum_{V\subseteq Z \land V \neq \varnothing } f^V\,.
	\end{split}
\end{equation}
Taking $f^{X\!Y\!Z}(X,Y,Z)$ as an example, we identify 
\begin{equation}
	\begin{split}
		f^{X\!Y\!Z}(X,Y,Z) = &\sum_{V\subseteq X\cup Y\cup Z} f^V - \sum_{X\subseteq X\cup Y} f^V - \sum_{V\subseteq X\cup Z} f^V - \sum_{V\subseteq Y\cup Z} f^V \\ &+ \sum_{V\subseteq X} f^V + \sum_{V\subseteq Y} f^V + \sum_{V\subseteq Z} f^V  - f^\varnothing\\
		=  &f(X,Y,Z) - f(X,Y,z) - f(X,y,Z) - f(x,Y,Z) \\ &+ f(X,y,z) + f(x,Y,z) + f(x,y,Z)  - f(x,y,z)\,,
	\end{split}
\end{equation}
where we have used $\sum_{W\subseteq V} f^W = f^V$, which can be shown by induction \cite{kuo2010decompositions}. This generalizes to all terms in \Cref{app-eq:anchoredf}.
Thus, we can rewrite the decomposition in \Cref{app-eq:anchored1} as
\begin{equation}
	\label{app-eq:anchored_featuresets}
	f(X,Y,Z) = \sum_{V\subseteq' X \cup Y \cup Z} f^V\,,
\end{equation}
with
\begin{equation}
	\label{app-eq:decomposition_terms_featuresets}
	f^V(V) = \sum_{W \subseteq' V } (-1)^{|V|-|W|} P_{\mathcal X \backslash W} f(X,Y,Z)\,, 
\end{equation}
where we defined $\subseteq'$ that does not break the feature sets $X,Y,Z$.

This decomposition generalizes in the obvious way beyond three sets to an arbitrary number of sets. There is only one distinguished point that qualifies as expansion point, namely the sample itself, i.e., $c=(x,y,z)$. In \Cref{app-sec:anchorpoint}, we discuss consequences of different choices for the anchor point. 
We now consider
\begin{equation}
	\begin{split}
		f(x,y,Z) &= f^\varnothing + f^Z(Z)\,\\
		f(x,Y,z) &= f^\varnothing + f^Y(Y)\,, \\
		f^{Y\!Z}(Y,Z) &= f(x,Y,Z) - f(x,Y,z) - f(x,y,Z) + f^\varnothing\,.
	\end{split}
\end{equation}
As relevances are not affected by constant factors, we have
\begin{equation}
	\label{app-eq:rel_sum}
	\begin{split}
		\bar{m}^{f^{Y}}_{Y|x} &= \bar{m}^{f|_{Z=z}}_{Y|x}\,\\
		\bar{m}^{f^{Z}}_{Z|x} &= \bar{m}^{f|_{Y=y}}_{Z|x}\,\\
		\bar{m}^{f^{Y\!Z}}_{Y\!Z|x} &= \bar{m}^{f}_{Y\!Z|x} - \bar{m}^{f|_{Z=z}}_{Y|x} - \bar{m}^{f|_{Y=y}}_{Z|x}\,,
	\end{split}
\end{equation}
This means that we can evaluate both main effects (first two rows) and second order effects (last row) by simple function evaluations without having to compute the full explicit decomposition in terms of original features. This holds for arbitrary feature sets $X$, $Y$ and $Z$.

\subsection{Choice of the anchor point}
\label{app-sec:anchorpoint}
In the previous section, we already picked the sample $(x,y,z)$ as anchor point. Here, we illustrate the impact of this choice and what consequences would arise from different choices. To identify how \textit{main effects} and \textit{joint effects} are distributed among the three terms in \Cref{eq:rel_sum}, we consider them before setting $(c^x,c^y,c^z)=(x,y,z)$,
\begin{equation} \label{app-eq:anchorexample}
	\begin{split}
		\bar{m}^{f^Y}_{Y|x} =& f(c^x, y, c^y) - \int f(c^x, Y, c^z) p(Y|x) dY\,, \\
		\bar{m}^{f^Z}_{Z|x} =& f(c^x, c^y, z)- \int f(c^x, c^y, Z) p(Z|x) dZ\, \\
		\bar{m}^{f_{Y\!Z}}_{Y\!Z|x} =& f(c^x,y,z) - \int f(c^x, Y, Z) p(Y,Z|x) dY dZ - \bar{m}^{f^Y}_{Y|x} - \bar{m}^{f^Z}_{Z|x}\,.
	\end{split}
\end{equation}	
We note that \textit{main effects} and \textit{joint effects} are shifted between the terms upon varying the anchor point of the decomposition. We demonstrate this by evaluating them for $f(Y,Z)=aY+aZ+bYZ$ for independent features with $p(Y/Z)=\mathcal{N}(0,\sigma_{Y/Z})$, where we find
\begin{equation}
	\begin{split}
		\bar{m}^{f^Y}_{Y} =& z_i(a+b c^y)\,, \\
		\bar{m}^{f^Z}_{Z} =& y_i(a+b c^z)\, \\
		\bar{m}^{f_{Y\!Z}}_{Y\!Z} =& b(y z-c^z y - c^y z)\,.
	\end{split}
\end{equation} 
This illustrates that the expansion point $(c^y,c^z)$ allows shifting relevances between main effects and joint effects. This is a well-known effect that has been observed already in linear models with multiplicative interactions, see for example the discussion in \cite{lengerich2020purifying}. Here, we argue that fixing the expansion point to the sample itself, i.e., $(c^x,c^y)=(x,y)$ in the example from above, is the only consistent choice in the \pd{} formalism for the following reasons:
\begin{enumerate}
	\item A different evaluation point than the sample itself is inconsistent with the original definition of \pd{} relevances in the sense that the property $\bar{m}^{f^Y}_{Y|x}=\bar{m}^f_{Y\!|x z}$ in case $p(Y|x)=p(Y|x,z)$ no longer holds.
	\item A different evaluation point than the sample itself will require to evaluate the model off the data manifold. This is exemplified in \Cref{app-eq:anchorexample}, where the first summand is in general not contained in the data manifold. Note that the integral in the second summand involves a conditional probability is not conditioned on $z$. This might still lead to an off-manifold evaluation in case of strongly correlated features, which is however inevitable.
	\item There is no other distinguished evaluation point apart from the sample itself. A different choice would require to impose a condition at the sample or the global level necessitating additional optimization procedures that would most likely turn the approach impractical for real-world applications.
\end{enumerate}

\subsection{Three- and $n$-point interactions} \label{app-sec:npoint_interaction}
Turning to three point interactions, we consider four feature sets $X=\{X^1,\ldots,X^k\}$, $A=\{A^1,\ldots,A^l\}$, $B=\{B^1,\ldots,B^m\}$ and $C=\{C^1,\ldots,C^n\}$ and an anchor point $c=(x^1,\ldots,x^k,a^1,\ldots,a^l,b^1,\ldots,b^m,c^1,\ldots,c^n)$. Analogously to the case of three sets, we can decompose an arbitrary function following \Cref{app-eq:anchored_featuresets} (already evaluating at the anchor point $X=x$ for simplicity):
\begin{equation}
	\begin{split}
		f(x,A,B,C) = f^\varnothing &+ f^{A}(A) + f^{B}(B) + f^{C}(C)\\ &+ f^{A\!B}(A,B) + f^{BC}(B,C) + f^{AC}(A,C) + f^{ABC}(A,B,C)\,.\\
	\end{split}
\end{equation}
The terms of this decomposition as given by \Cref{app-eq:decomposition_terms_featuresets} read
\begin{equation}
	\begin{split}
		f^A(A) &= f(x,A,b,c) - f_\varnothing\,,\\
		f^B(B) &= f(x,a,B,c) - f_\varnothing\,,\\
		f^C(C) &= f(x,a,b,C) - f_\varnothing\,,\\
		f^{AB}(A,B) &= f(x,A,B,c) - f(x,A,b,c) - f(x,a,B,c) + f^\varnothing\,,\\
		f^{BC}(B,C) &= f(x,a,B,C) - f(x,a,B,c) - f(x,a,b,C) + f^\varnothing\,,\\
		f^{AC}(A,C) &= f(x,A,b,C) - f(x,A,b,c) - f(x,a,b,C) + f^\varnothing\,,\\
		f^{ABC}(A,B,C) &= f(x,A,B,C) -f(x,A,b,C) -f(x,A,B,c) -f(x,a,B,C)\\
		&+f(x,A,b,c)+f(x,a,B,c)+f(x,a,b,C) -f^\varnothing\,.
	\end{split}
\end{equation}
This translates into the following expressions for the (interaction) relevances that can be evaluated as efficiently as in the case of the two-point interactions above,
\begin{equation}
	\begin{split}
		\bar{m}^{f^{A}}_{A|x} &= \bar{m}^{f|_{B=b,C=c}}_{A|x}\,\\
		\bar{m}^{f^{B}}_{B|x} &= \bar{m}^{f|_{A=a,C=c}}_{B|x}\,\\
		\bar{m}^{f^{C}}_{C|x} &= \bar{m}^{f|_{A=a,B=b}}_{C|x}\,\\
		\bar{m}^{f^{AB}}_{AB|x} &= \bar{m}^{f|_{C=c}}_{AB|x}- \bar{m}^{f|_{B=b,C=c}}_{A|x} - \bar{m}^{f|_{A=a,C=c}}_{B|x} \,\\
		\bar{m}^{f^{BC}}_{BC|x} &= \bar{m}^{f|_{A=a}}_{BC|x}- \bar{m}^{f|_{A=a,C=c}}_{B|x} - \bar{m}^{f|_{A=a,B=b}}_{C|x} \,\\
		\bar{m}^{f^{AC}}_{AC|x} &= \bar{m}^{f|_{B=b}}_{AC|x}- \bar{m}^{f|_{B=b,C=c}}_{A|x} - \bar{m}^{f|_{A=a,B=b}}_{C|x} \,\\
		\bar{m}^{f^{ABC}}_{ABC|x} &= \bar{m}^{f}_{ABC|x} - \bar{m}^{f^{AB}}_{AB|x} - \bar{m}^{f^{BC}}_{BC|x} - \bar{m}^{f^{AC}}_{AC|x} - \bar{m}^{f^{A}}_{A|x} - \bar{m}^{f^{B}}_{B|x} - \bar{m}^{f^{C}}_{C|x}\,.
	\end{split}
\end{equation}
Also this scheme generalizes in the same manner to interactions between an arbitrary number of sets.
\subsection{Implication for relevance decompositions} \label{app-sec:npoint_completeness_relation}
Given the general decomposition of the form discussed in the previous section, if we want to study the interaction between two feature \emph{sets} $Y$ and $Z$ given the remaining features $X$, we can write 
\begin{equation}
	\begin{split}
		f(X,Y,Z) = f^\varnothing &+ f^X(X) + f^Y(Y) +f^Z(Z) \\ &+f^{X\!Y}(X,Y) + f^{X\!Z}(X,Z) + f^{Y\!Z}(Y,Z) + f^{X\!Y\!Z}(X,Y,Z)\,
	\end{split}
\end{equation}
where all functions on the right hand side implicitly depend on the anchor point $(x,y,z)$. By construction of the decomposition all terms on the right hand side that involve an $x$ as subscript vanish if we evaluate at $X=x$, hence 
\begin{equation}
	f(x,Y,Z) = f^\varnothing + f^Y(Y) +f^Z(Z) + f^{Y\!Z}(Y,Z)
\end{equation}
We can now compute the relevance of the function on the left hand side
\begin{equation}
	\bar{m}^f_{Y\!Z|x} = \bar{m}^{f^Y}_{Y|x} + \bar{m}^{f^Z}_{Z|x} +\bar{m}^{f^{Y\!Z}}_{Y\!Z|x}\,,
\end{equation}
where we used that $\bar{m}^{f^\varnothing}_{\varnothing|x}=0$ and $\bar{m}^{f^Y}_{Y|x}=\bar{m}^{f^Y}_{Y\!Z|x}$. We refer to the first two terms as \emph{main effects} and to the third term as \emph{joint effects} between the two sets $Y$ and $Z$, see \Cref{sec:interpretation} for details. 

Turning to interactions between three feature sets $A$, $B$, $C$ in dependence of the remaining features $X$, we have the following decomposition at $X=x$
\begin{equation}
	\begin{split}
		f(x,A,B,C) = f^\varnothing &+ f^A(A) + f^B(B) +f^C(C)\\ &+ f^{AB}(A,B) + f^{BC}(B,C) + f^{AC}(A,C) + f^{ABC}(A,B,C)\,.
	\end{split}
\end{equation}
Computing relevances yields the \emph{interaction completeness} property
\begin{equation}	\label{app-eq:completeness_threepoint}
	\begin{split}
		\bar{m}^f_{ABC|x} =& \bar{m}^{f^A}_{A|x} + \bar{m}^{f^B}_{B|x} + \bar{m}^{f^C}_{C|x}\\
		&+\bar{m}^{f^{AB}}_{AB|x}+\bar{m}^{f^{BC}}_{BC|x}+\bar{m}^{f^{AC}}_{AC|x}\\
		&+\bar{m}^{f^{ABC}}_{ABC|x}\,,
	\end{split}
\end{equation}
where we refer to the terms in the first line as \emph{main effects}, to the terms in the second line as \emph{second order joint effects} and to the terms in the third line as \emph{third order joint effects}. This notion generalizes to interactions between an arbitrary number of sets and efficient projections exist for all of these terms, see previous section.

\subsection{Shielded decomposition for three-point interactions}
\label{app-sec:shielded3pt}
We arrive at the fully shielded main effects for $A$ by first imagining $B$ and $C$ to be a single (combined) feature set $\{BC\}$ and proceeding as before,
\begin{equation}
	\begin{split}
		f^{A\backslash BC} &= f^A + f^{A \{BC\}} = f^A + f^{ABC} + f^{AB} + f^{BC}	\\
		&= f(x, A, B, C) - f(x, a, B, C)\,,
	\end{split} 
\end{equation}
which results in 
\begin{equation}
	\bar{m}^{f^{A}}_{A\backslash BC|x} = \bar{m}^{f(x, A,B,C)}_{ABC|x} - \bar{m}^{f(x, a,B,C)}_{BC|x}\,.
\end{equation}
Again this term has a direct Shapely counterpart. Similarly one can define the contribution of $AB$ shielded from $C$
\begin{equation}
	\begin{split}
		f^{AB\backslash C} &= f^{\{AB\}} + f^{\{AB\}C} = f^A + f^B + f^{AB} +f^{ABC} + f^{AC} + f^{BC}	\\
		&= f(x, A, B, C) - f(x, a, b, C)\,.
	\end{split} 
\end{equation}
Eventually, this leads to the following \emph{interaction completeness} relation in terms of shielded effects
\begin{equation}
	\begin{split}
		\bar{m}^f_{ABC|x} =& -\bar{m}^{f^{A}}_{A\backslash BC|x}-\bar{m}^{f^{B}}_{B\backslash AC|x}-\bar{m}^{f^{C}}_{C\backslash AB|x}\\
		&+\bar{m}^{f^{AB}}_{AB\backslash C|x}+\bar{m}^{f^{BC}}_{BC\backslash A|x}+\bar{m}^{f^{AC}}_{AC\backslash B|x}\\
		&+\bar{m}^{f^{ABC}}_{ABC|x}\,,
	\end{split}
\end{equation}
which generalizes \Cref{eq:completenessshielded}.

\section{Interaction relevance for classification}
\label{app-sec:derivationinteraction}
We consider (non-overlapping) feature sets $X$, $Y$, $Z$ that cover the set of all features and denote the class label by $c$. In the following, we look at a generalization of \textit{informative conditional interactions} \cite{jakulin2003quantifying,henelius2017interpreting}, i.e., interactions that satisfy $p(Y,Z|c)=p(Y|c)p(Z|c)$. Here, we consider a slight generalization, where $Y$ and $Z$ are assumed to be conditionally independent given $c$ and $x$, 
\begin{equation}
	\label{app-eq:definformativecond2}
	p(Y, Z|c,x)= p(Y|c,x) p(Z|c,x)\,,
\end{equation}
or equivalently
\begin{equation}
	\label{app-eq:definformativecond}
	p(y,z,x|c)= \frac{p(x,y|c) p(x,z|c)}{p(x|c)}\,.
\end{equation}
Using Bayes' rule, we obtain for the class probability given the features
\begin{equation} \label{app-eq:bayes_condindependence}
	p(c|x,y,z)=\frac{p(c) p(x,y|c)p(x,z|c)}{p(x|c) p(x,y,z)}\,.
\end{equation}

\noindent
The   centered $m$-values are then given by
\begin{align}
	\bar{m}^{f_c}_{Y\!Z|x} &= \log_2\left(\frac{p(c|x,y,z)}{p(c|x)}\right) \,,\label{app-eq:mf1} \\
	\bar{m}^{f_c^Y}_{Y\!Z|x} &= \log_2\left(\frac{p(c|x,y,z)}{ \int \text{d}Y p(c|x, Y, z) p(Y|x)}\right) \, ,\\ 
	\bar{m}^{f_c^Z}_{Y\!Z|x} &= \log_2\left(\frac{p(c|x,y,z)}{ \int \text{d}Z p(c|x, y, Z) p(Z|x)}\right) \,.\label{app-eq:mf3}
\end{align}
From this we derive the joint effect of $Y$ and $Z$ without any assumption on imputer distribution
\begin{align} \label{app-eq:clf_nointeractionresult}
	\begin{split}
		\bar{m}^{f_{c}^{Y\!Z}}_{Y\!Z|x} &= \bar{m}^{f_c}_{YZ|x} - \bar{m}^{f_c^{Z}}_{YZ|x} - \bar{m}^{f_c^{Y}}_{YZ|x}  \\
		&= \log_2\left(\frac{p(y,z|x)}{p(y|x)p(z|x)}\right) + \log_2\left(\frac{1}{p(x|c)}\int \text{d}Y\text{d}Z p(x, Y, Z|c)\frac{p(Y|x)}{p(Y|x,z)}\frac{p(Z|x)}{p(Z|x,y)}\right)\,,
	\end{split}
\end{align}
where we leverage the \emph{no-interaction} property via \Cref{app-eq:bayes_condindependence}.
Thus, the interaction relevance vanishes identically, if we additionally ensure $y, z$ to be conditionally independent given the remaining features $x$, i.e.,
\begin{equation}
	q(y,z|x) = q(y|x) q(z|x)\,.
	\label{app-eq:defdatacondition}
\end{equation}
We show in \Cref{sec:classification} and \Cref{app-sec:sampleapproximation} how this condition on the imputer distribution can be handled in practice. 

The vanishing interaction relevance under the given conditions relies on using logarithmic differences in the relevance definition in \Cref{eq:defmhat}. In particular, \Cref{app-eq:mf1}-(\ref{app-eq:mf3}) are tied to this choice. 
Using centered relevances defined via differences of raw probabilities or log odds, as two other popular choices in the literature, would lead to a violation of the no-interaction property. 
Hence, the analysis of interaction effects singles out logarithmic differences as relevance measure among the three most popular relevance measures. This is analogue to the analysis related to Shapely values, see \Cref{sec:shapley} and \Cref{app-sec:shapley} .

\subsection*{\emph{Complete} conditioning and the mutual information dilemma}
In \Cref{app-eq:clf_nointeractionresult} the second terms dependence on the classifier and is a consequence of ignoring the $y$/$z$ dependence in when evaluating the joint effect of both features. It is insightful to consider what happens if one would use the \emph{correct}, as judged by the classifier, conditioning. This would correspond to the classification value function \Cref{eq:defvaluefct_classification} for Shapley values. The centered $m$-values are then given by
\begin{align}
	\bar{m}^{f_c}_{Y|xz}&=\log_2\left(\frac{p(c|x,y,z)}{p(c|x,z)}\right)\,,\\
	\bar{m}^{f_c}_{Z|xy}&=\log_2\left(\frac{p(c|x,y,z)}{p(c|x,y)}\right)\,,\\
	\bar{m}^{f_c}_{Y\!Z|x}&=\log_2\left(\frac{p(c|x,y,z)}{p(c|x)}\right)\,,
\end{align}
from which the \pd{} joint effect follows to be
\begin{align}
	\bar{m}^{f_{c}^{Y\!Z}}_{Y\!Z|x} 
	&\approx \bar{m}^{f_c}_{y|xz}+\bar{m}^{f_c}_{z|xy}-\bar{m}^{f_c}_{yz|x}\notag\\
	& \approx \log_2\left(\frac{p(z|x)p(y|x)}{p(y,z|x)}\right) =: i(y : z|x).
\end{align}
This term is conventionally referred to as \emph{local conditional mutual information}. This local mutual information is closely related to the mutual information via
\begin{equation}
	I(Y : Z|x) = \mathbb{E}_{Y, Z}\big[i(Y : Z| x)\big]\,.
\end{equation}
It measures the joint information content of $y$ and $z$ and vanishes if they are independent. 
Importantly, we cannot simplify everything through specializing to cases for which the local mutual information vanishes, e.g., conditional independent data distributions.
\begin{equation}
	p(y,z|x) \stackrel{\text{(\ref{app-eq:definformativecond2})}}{=} \int \text{d}c \,\, p(y|c,x) p(z|c,x) p(c|x) \stackrel{\text{!}}{=}p(y|x)p(z|x) \,.
\end{equation}
The simplest way of achieving this is via $p(y|c,x)=p(y|x)$ or equivalently, $p(y,c|x)=p(y|x)p(c|x)$ (requiring this for either $y$ or $z$ is sufficient). However, this renders either $y$ or $z$ uninformative for the prediction. We dub this the local \emph{mutual information dilemma}. It states that we either have to explicitly calculate the local mutual information, which is difficult in practice, or alternatively, break the feature dependencies and thereby inevitably evaluate the model off-manifold.

\section{AND, OR, XOR regression examples} \label{app-sec:and_or_xor}
\begin{table}[h!]
	\caption{\pd{} raw main and joint effects for $X \wedge Y$, $=X \vee Y$ and $X \veebar Y$ and a uniform data distribution (up to a constant $^1/_4$).}
	\begin{center}
		\begin{tabular}{l|cccc|cccc|cccc}
			\toprule
			$f(X,Y)$ & \multicolumn{4}{c|}{$X \wedge Y$} & \multicolumn{4}{c|}{ $X \vee Y$} & \multicolumn{4}{c}{$X \veebar Y$} \\
			$(x,y)$ & (0, 0) & (0, 1) & (1, 0) & (1, 1) & (0, 0) & (0, 1) & (1, 0) & (1, 1) & (0, 0) & (0, 1) & (1, 0) & (1, 1) \\
			\midrule
			$\bar m^{f^X}_X$       &      0 &     -2 &      0 &      2 &     -2 &      0 &      2 &      0 &     -2 &      2 &      2 &     -2 \\
			$\bar m^{f^Y}_Y$       &      0 &      0 &     -2 &      2 &     -2 &      2 &      0 &      0 &     -2 &      2 &      2 &     -2 \\
			$\bar m^{f^{X\!Y}}_{X\!Y}$ &     -1 &      1 &      1 &     -1 &      1 &     -1 &     -1 &      1 &      2 &     -2 &     -2 &      2 \\
			\bottomrule
		\end{tabular}
	\end{center}
	\label{app-tab:and_or_xor_raw}
\end{table}	

\begin{table}[h!]
	\caption{\pd{} shielded main and joint effects for $X \wedge Y$, $=X \vee Y$ and $X \veebar Y$ and a uniform data distribution (up to a constant $^1/_4$).}
	\begin{center}
		\begin{tabular}{l|cccc|cccc|cccc}
			\toprule
			$f(X,Y)$ & \multicolumn{4}{c|}{$X \wedge Y$} & \multicolumn{4}{c|}{ $X \vee Y$} & \multicolumn{4}{c}{$X \veebar Y$} \\
			$(x,y)$ & (0, 0) & (0, 1) & (1, 0) & (1, 1) & (0, 0) & (0, 1) & (1, 0) & (1, 1) & (0, 0) & (0, 1) & (1, 0) & (1, 1) \\
			\midrule
			$\bar m^{f^X}_{X\backslash Y}$    &     -1 &     -1 &      1 &      1 &     -1 &     -1 &      1 &      1 &      0 &      0 &      0 &      0 \\
			$\bar m^{f^Y}_{Y\backslash X}$    &     -1 &      1 &     -1 &      1 &     -1 &      1 &     -1 &      1 &      0 &      0 &      0 &      0 \\
			$\bar m^{f^{X\!Y}}_{\backslash X\!Y}$ &      1 &     -1 &     -1 &      1 &     -1 &      1 &      1 &     -1 &     -2 &      2 &      2 &     -2 \\
			\bottomrule
		\end{tabular}
	\end{center}
	\label{app-tab:and_or_xor_shielded}
\end{table}	
We consider two binary input variables $X$ and $Y$ that are sampled uniformly, i.e., are subject to the data distribution $p(X,Y)=\frac{1}{4}$. For the three functions $f(X,Y)= X \wedge Y$, $g(X,Y)= X \vee Y$, $h(X,Y)= X \veebar Y$, we work out the raw and shielded \pd{} effects in \Cref{app-tab:and_or_xor_raw} and \Cref{app-tab:and_or_xor_shielded}. 

Because $X \vee Y$, $X \wedge Y$ and $X \veebar Y$ share the same shielded joined effects up to a constant factor and the shielded main effects vanish for $X \veebar Y$, we can understand $X \vee Y$ and $X \wedge Y$ are versions of $X \veebar Y$ modified with main effects, as already demonstrated in \cite{lengerich2020purifying}. This result is slightly unintuive at first and illustrates the danger of inferring intuitive ground truth relevances and interactions for seemingly simple functions.

\clearpage
\section{Additional plots: synthetic dataset} 
\label{app-sec:syntheticdataset}
\begin{figure}[h!]
	\centering
	\begin{tabular}{c c}
		\small{(a) \pd{} relevances}	&  \small{(b) shielded main effects}	\\
		\includegraphics[width=.4\linewidth]{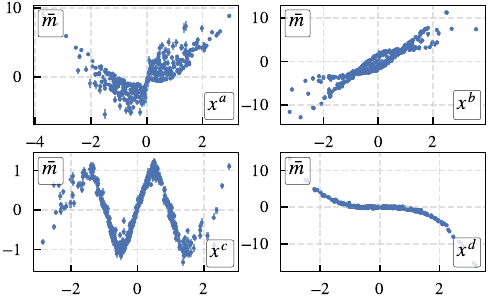} &	
		\includegraphics[width=.4\linewidth]{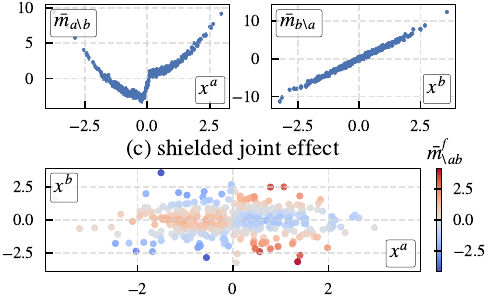} \\
	\end{tabular}
	\caption{Fully-Connected Network analyzed with \pd{} for synthetic regression task. 
		(a):~~\pd{} relevances, \Cref{eq:defmbar}. 
		(b): shielded main effects,  \Cref{eq:completenessshielded}. 
		(c): color-encoded shielded joint effect of feature $X_a$ and $X_b$. Interaction is given by $\text{sgn}(X_a)\,|X_b|$. }
	\label{app-fig:fcnn_synthetictask}
\end{figure}

\begin{figure}[h!]
	\centering
	\begin{tabular}{c c}
		\small{(a) \pd{} relevances}	&  \small{(b) shielded main effects}	\\
		\includegraphics[width=.4\linewidth]{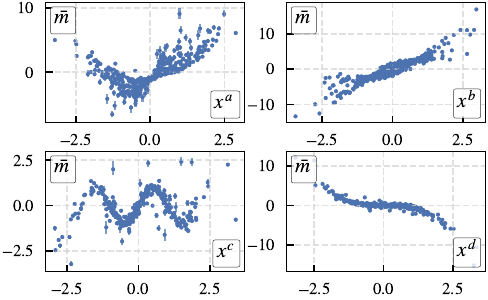} &	
		\includegraphics[width=.4\linewidth]{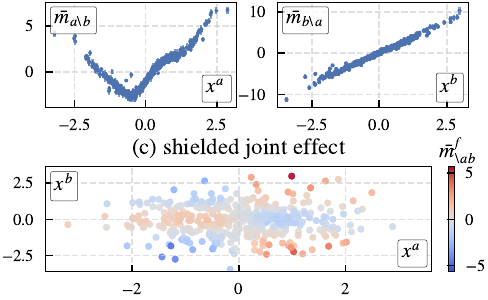} \\
	\end{tabular}
	\caption{Gaussian Process model analyzed with \pd{} for synthetic regression task. 
		(a):~~\pd{} relevances, \Cref{eq:defmbar}. 
		(b): shielded main effects,  \Cref{eq:completenessshielded}.
		(c): color-encoded shielded joint effect of feature $X_a$ and $X_b$. Interaction is given by $\text{sgn}(X_a)\,|X_b|$. }
		\label{app-fig:GP_synthetictask}
\end{figure}

For the readers convenience we present attributions for two alternative model categories: (i) a fully-connected neural network (\Cref{app-fig:fcnn_synthetictask}) and (ii) a gaussian process (\Cref{app-fig:GP_synthetictask}). 
\begin{figure}[h!]
	\centering
	\begin{tabular}{c c}
		\small{(a) shapley values}	&  \small{(b) shapley main effects}	\\
		\includegraphics[width=.4\linewidth]{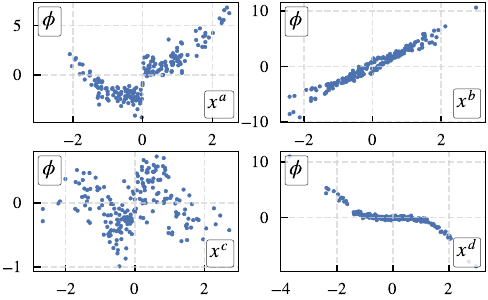} &	
		\includegraphics[width=.4\linewidth]{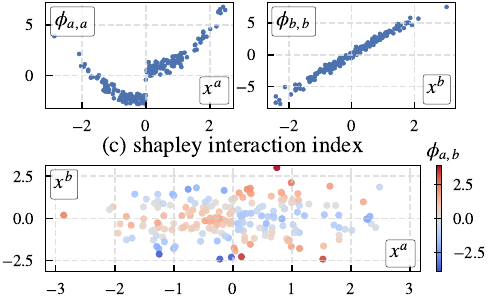} \\
	\end{tabular} 
	\caption{Shapley values for a random forest applied on synthetic regression tasks using $\#=400$ model calls. 
		(a)~~Shapley values (b)~~Shapley main effects (c)~~Shapley Interaction Index associated with $X_a$ and $X_b$. }
	\label{app-fig:custom_shapley_synthetic}
\end{figure}

\newpage

\subsection{Relevances in the presence of correlated features}
\label{app-sec:correlation}

\begin{figure}[h!]
	\centering
	\begin{tabular}{c c}
		\small{(a) \pd{} relevances}	&  \small{(b) shielded main effects}	\\
		\includegraphics[width=.4\linewidth]{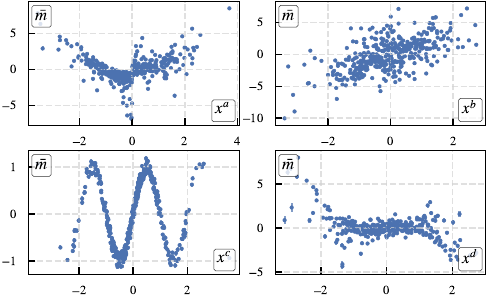} &	
		\includegraphics[width=.4\linewidth]{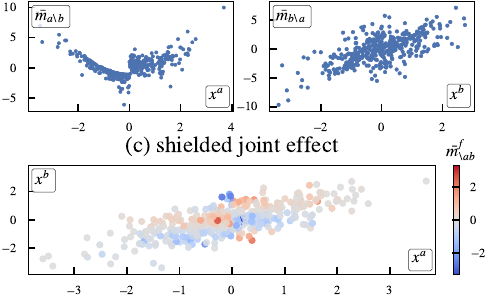} \\
	\end{tabular} 
	\caption{Analyzing the analytic function \Cref{eq:targetfunction} with correlated features ($\rho=0.7$) using \pd{} ($\#=800$).  
		(a):~~\pd{} relevances. 
		(b): shielded main effects. 
		(c): shielded joint effect of feature $X_a$ and $X_b$.}
	\label{app-fig:correlated_preddiff}
\end{figure}

\begin{figure}[h!]
	\centering
	\begin{tabular}{c c}
		\small{(a) shapley values}	&  \small{(b) shapley main effects}	\\
		\includegraphics[width=.4\linewidth]{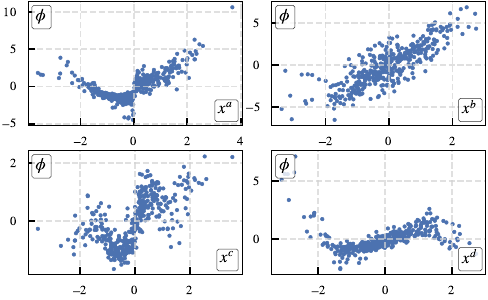} &	
		\includegraphics[width=.4\linewidth]{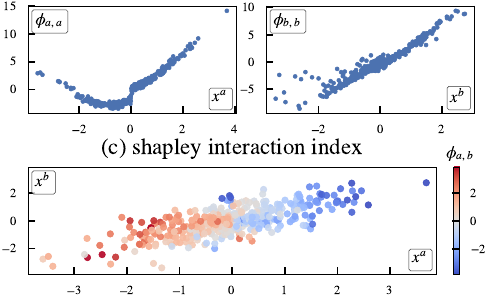} \\
	\end{tabular}
	\caption{Analyzing the analytic function \Cref{eq:targetfunction} with correlated features ($\rho=0.7$) using Shapley values ($\#=800$).  
	(a)~~Shapley values (b)~~Shapley main effects (c)~~Shapley Interaction Index associated with $X_a$ and $X_b$.  }
	\label{app-fig:correlated_shapley}
\end{figure}

We repeat the synthetic regression task with correlated features with unit variance and correlation $\rho=0.7$. To avoid ambiguities due to model training, we directly use the analytic function \Cref{eq:targetfunction}. 
For correlated features, \pd{} and Shapley values show a qualitatively different behavior. In the limit of perfectly correlated features, their attributions are ambiguous without additional causal assumptions.
In this setting, \pd{} single-feature attributions tend towards zero as $m^{f_c}_{Y|x}\to f_c(x,y)$ if $Y$ denotes one of the correlated features in question. In contrast, Shapley values distribute relevance evenly across all features. 
This can be seen as a sign for a higher reliability of \pd{} relevances, since positive/negative attributions are guaranteed to be caused by the model. In this sense \pd{} is true to the model and true to the data. 
In \Cref{app-fig:correlated_shapley} the Shapley values for $X_c$ and $X_d$ are tilted in comparison to the uncorrelated setting in \Cref{app-fig:custom_shapley_synthetic}. For the Shapley Interaction Index, the same effect occludes the true interaction.
In contrast, \pd{} attributions in \Cref{app-fig:correlated_preddiff} are structurally equivalent to the independent feature setting, i.e., the functional form of sine and interaction are still clearly recognizable. However, this comes at the price of partially less pronounced attributions.

 \subsection{Comparing convergence using a white box regressor: \pd{} vs. Shapley values}
 \label{sec:dynamask}
 In the following, we build on \cite{crabbe2021explaining, ismail2020benchmarking} to investigate the convergence of \pd{} vs. Shapley attributions based on known ground-truth attributions. 
The task is to recover the relevant features (interactions) given the attributions of a white box regression model.
 The analysis is based on 25 Gaussian features $\mathcal{X} = \{X_1, \ldots X_{25}\}$ with unit variance and a correlation of $\rho=0.3$.  The target function is based on 10 randomly selected features $\mathcal{A} = \{i_1, \ldots, i_{10}\}$ and 20 random feature pairs $\mathcal{B}=\{(i_1, j_1), \ldots (i_{20}, j_{20})\}$ with $i, j\in \mathcal{A}$.
We focus on interactions between elements in $\mathcal{A}$, since arbitrary interactions generally induce corresponding main effects, see \cite{lengerich2020purifying}.
The overall target function is given by 
 \begin{equation} \label{eq:whitebox_regressor}
 	f(\mathcal{X}) = \sum_{i \in \mathcal{A}} (X_i)^2 + \sum_{(i, j) \in \mathcal{B}} (X_i X_{j}) \,.
 \end{equation}
This task singles out a well-defined ground-truth, both on the level of relevances ($\mathcal{A}$) and pairwise interactions ($\mathcal{B}$), i.e., binary masks on features and pairs of features. 
To obtain a more challenging task, we include additive white noise based on all non-contributing pairwise features $\bar{\mathcal{B}}$ with variance $\epsilon = 0.01$.

Within this setup we compute attributions for 200 random samples. 
The absolute value of main and interaction relevances are compared to the respective ground truth. As in \cite{ismail2020benchmarking}, we base our analysis on precision and recall to assess whether all features identified as salient were in fact informative (precision) and whether all informative features were identified (recall), over a range of thresholds. 
To summarize the behavior through a single number, we chose the average precision score\footnote{We leverage the implementation in scikit-learn.}, which quantifies the area under the precision recall curve, and for completeness also state the AUC-ROC score. 
We repeat this experiment three times to obtain error estimates. The results for varying computational costs (i.e., numbers of function evaluations) are summarized in \Cref{tab:convergence_dynamask}. Since this model is inherently additive, \pd{} main effects perfectly recover the relevant features with minimal computational effort. 
In contrast, Shapley values need to sample many coalitions to reveal the simple underlying structure. 
Revealing the sparse interactive structure is challenging for both methods. However, \pd{} consistently outperforms Shapley values independent of the number of model calls~$\#$.

\begin{table}
	\caption{Convergence analysis based on the white-box regressor \Cref{eq:whitebox_regressor}. We provide AUC-ROC and average precision scores and state only statistical significant digits. Compared to Shapley values, \pd{} is numerically inexpensive and more capable of recovering the ground-truth both in terms of relevance and interactions.  }
	\begin{tabular}{llccccc}
		\toprule
					&  		  & 			\multicolumn{2}{c}{$\# = 10$ \hspace{1mm}}							&&					\multicolumn{2}{c}{$\#=600$}							\\ 
					&												& 	AUC-ROC		& 	avg.~Prec.		&&	AUC-ROC	& avg.~Prec. 		\\ \cmidrule{3-4} \cmidrule{6-7}
		Main effects	   & \pd{}		&			0.915				&	$0.910$	 	&& 0.925	&			$0.918$		\\
									 & Shapley		&				0.567				&	$0.473$	  	&&	0.801	&			$0.755$			\\ \midrule
		Interactions		& \pd{}			&			0.726				&	$0.279$		&& 0.717	&		$0.311$		\\	
									 & Shapley			&			0.496				&	$0.066$		&&	0.564	&	$0.087$		\\
		\bottomrule
	\end{tabular}
\label{tab:convergence_dynamask}
\end{table}

\newpage

\section{Additional plots: NHANES}
\label{app-sec:nhanes}
\begin{figure}[ht]
	\centering
	\begin{tabular}{ccc}
		&  (a) Relevances	& \\
		\includegraphics[width=0.33\textwidth,align=c]{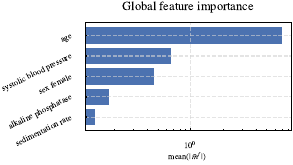} &
		\includegraphics[width=0.3\textwidth,align=c]{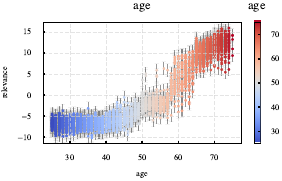}	&
		\includegraphics[width=0.3\textwidth,align=c]{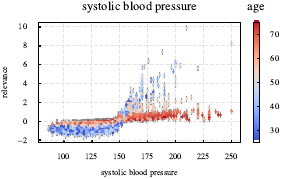}	\\ \noalign{\smallskip}
		&  (b) Interactions & \\
		\includegraphics[width=0.33\textwidth,align=c]{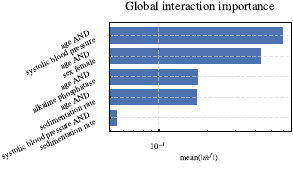} &
		\includegraphics[width=0.3\textwidth,align=c]{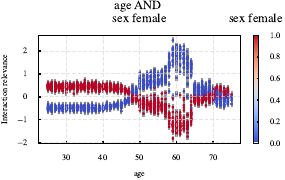}	&
		\includegraphics[width=0.3\textwidth,align=c]{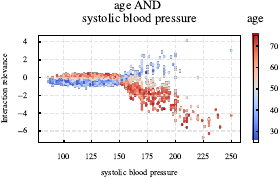}
	\end{tabular}

	\caption{\pd{} (interaction) relevances for a \textit{Random Forest} trained on the NHANES dataset using a (marginal) train set imputer \cite{aas2021explaining} that are in qualitative agreement with existing methods \cite{lundberg2020local}. 
		top left:~~Ranking of the five most important features. 
		top center:~~Relevance for the most important feature age.
		top right:~~Relevance for the second most important feature systolic blood pressure.
		bottom left:~~Ranking of the five most important feature interactions.
		bottom center:~~Interaction relevance between systolic blood pressure and age.
		bottom right:~~Interaction relevance between age and sex revealing a pronounced age dependence.}
	\label{app-fig:nhanes_relevances_ts}
\end{figure}

\Cref{app-fig:nhanes_relevances_ts} shows results for the (marginal) train set imputer.

\newpage 
\section{Additional plots: MNIST}
\label{app-sec:mnist}
\begin{figure}[h!]
	\centering
	\begin{tabular}{l || r}
		\includegraphics[width=.34\columnwidth]{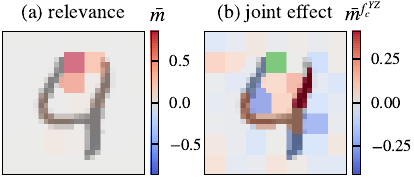}  &
		\includegraphics[width=.34\columnwidth]{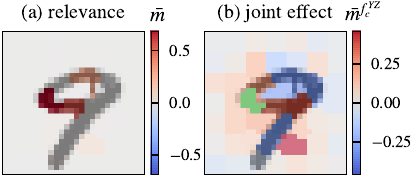}\\ 
		\includegraphics[width=.34\columnwidth]{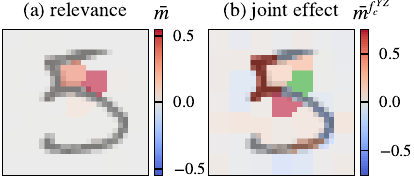} &
		\includegraphics[width=.34\columnwidth]{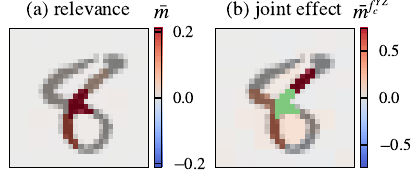} 	
	\end{tabular}
	\caption{\pd{} for MNIST digits calculated on $\sim50$ SLIC superpixels \cite{achanta2012slic} using a (marginal) train set imputer for the true (and correctly predicted) class label. (a)~~\pd{} relevances/main effects, \Cref{eq:defmhat} (b)~~\pd{} joint effects, \Cref{eq:rel_sum} with respect to the marked (green) reference super-pixel of highest relevance. 
	We used $\#=600$ imputations.}
	\label{app-fig:mnist}
\end{figure}

\begin{figure}[h!]
	\centering
	\begin{tabular}{l || r}
		\includegraphics[width=.34\columnwidth]{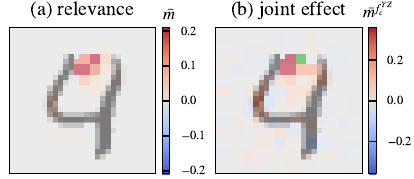}  &
		\includegraphics[width=.34\columnwidth]{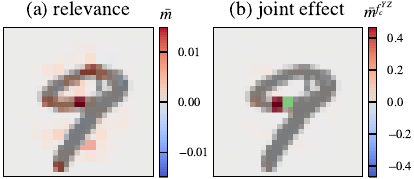}\\ 
		\includegraphics[width=.34\columnwidth]{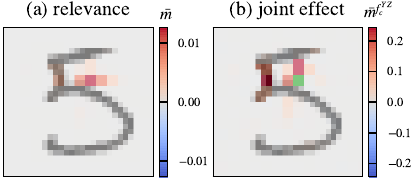} &
		\includegraphics[width=.34\columnwidth]{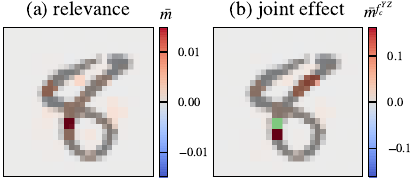} 	
	\end{tabular}
	\caption{\pd{} for MNIST digits calculated on $\sim150$ SLIC superpixels \cite{achanta2012slic} using a (marginal) train set imputer for the true (and correctly predicted) class label. (a)~~\pd{} relevances/main effects, \Cref{eq:defmhat} (b)~~\pd{} joint effects, \Cref{eq:rel_sum} with respect to the marked (green) reference super-pixel of highest relevance. }
	\label{app-fig:mnist_trainset_150}
\end{figure}

\begin{figure}[h!]
	\centering
	\begin{tabular}{l || r}
		\includegraphics[width=.34\columnwidth]{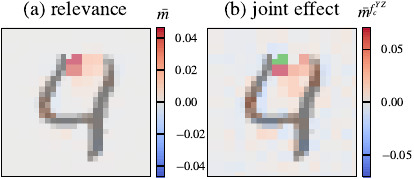}  &
		\includegraphics[width=.34\columnwidth]{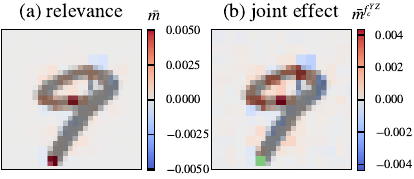}\\ 
		\includegraphics[width=.34\columnwidth]{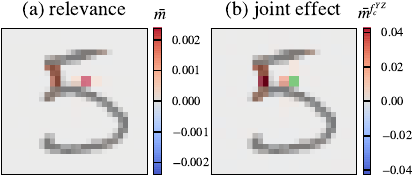} &
		\includegraphics[width=.34\columnwidth]{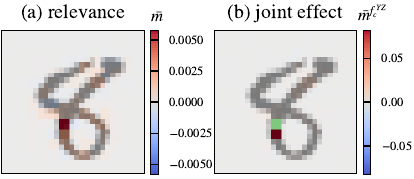} 	
	\end{tabular}
	\caption{\pd{} for MNIST digits calculated on $\sim150$ SLIC superpixels \cite{achanta2012slic} using a (conditional) vae imputer for the true (and correctly predicted) class label. (a)~~\pd{} relevances/main effects, \Cref{eq:defmhat} (b)~~\pd{} joint effects, \Cref{eq:rel_sum} with respect to the marked (green) reference super-pixel of highest relevance. }
	\label{app-fig:mnist_vae_150}
\end{figure}

\begin{figure}[h!]
	\centering
	\begin{tabular}[t]{l| c | c}
	\multicolumn{1}{c}{}	&  (a) Relevances		&			(b) Interactions	\\
	\pd{} & 	\includegraphics[width=.4\columnwidth]{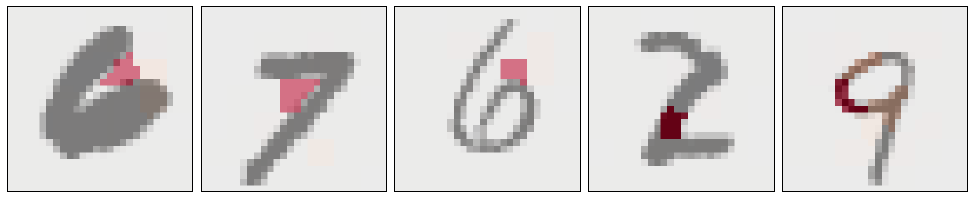}				&			\includegraphics[width=.4\columnwidth]{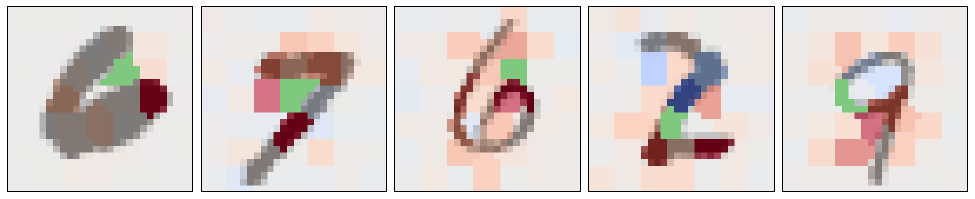}			\\ 
	Shapley	&		\includegraphics[width=.4\columnwidth]{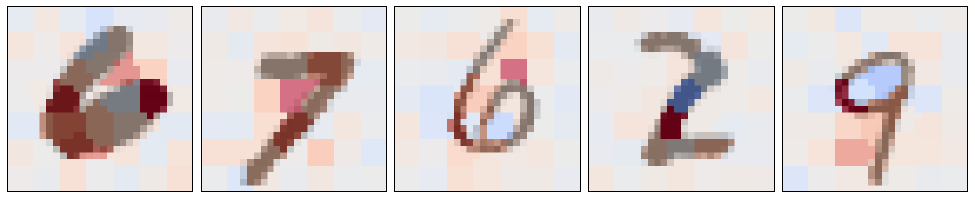}				&				\includegraphics[width=.4\columnwidth]{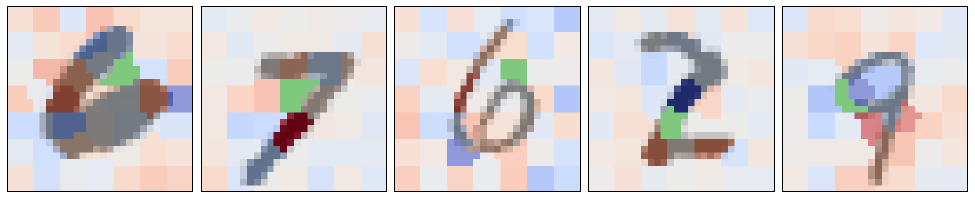}	\\
\end{tabular}
	\caption{
		Comparing \pd{}  relevances/joint effect and Shapley values/interaction index on randomly selected digits for the true (and correctly predicted) class label using a (marginal) train set imputer. Interaction measured with respect to the marked (green) reference super-pixel of highest relevance. \pd{} and Shapley values produce qualitatively similar feature and interaction attributions based on $\# = 600$ model calls.}
	\label{app-fig:mnist_shapcomparison_trainset}
\end{figure}

\begin{figure}[h!] 
	\centering
	\begin{tabular}[t]{c  c}
		  (marginal) Train Set 	&			(conditional) VAE\\
			\includegraphics[width=.4\columnwidth]{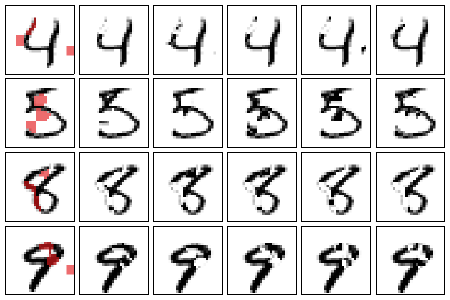}				&			\includegraphics[width=.4\columnwidth]{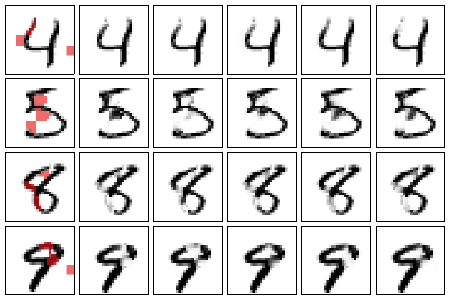}			\\ 
	\end{tabular}
	\caption{
		Imputed samples generated with a (marginal) Train Set imputer compared to a (conditional) VAE Imputer for three independent patches. Digits are identical to \Cref{fig:mnist}.} 
	\label{app-fig:imputations}
\end{figure}

We show attributions for a (marginal) train set imputer in \Cref{app-fig:mnist} and \Cref{app-fig:mnist_shapcomparison_trainset}. \pd{} relevance attributions are qualitatively very similar to the (conditional) VAE imputer in \Cref{fig:mnist} and \Cref{fig:mnist_shapcomparison}. 
Interactions, as measured by the joint effect, are also similar. However, the VAE joint effects are more pronounced and sparse, which makes them easier to interpret, e.g., consider digit four and nine in \Cref{fig:mnist} for which all \emph{background} attributions are removed.
In contrast, overall, the important, highly interacting superpixel do not change. 
Additionally, we show attributions for marginal and conditional imputer with more fine grained superpixels in \Cref{app-fig:mnist_trainset_150} and \Cref{app-fig:mnist_vae_150} respectively.
To further highlight the qualitative differences between both imputers, we show example imputations in \Cref{app-fig:imputations}. As expected, the VAE imputations are more realistic but consequently less diverse. This is the reason for their more targeted attributions. 


\section{Additional plots: CUB Birds} \label{app-sec:cubbirds}
\begin{figure}[h!]
	\centering
	(a) High relevance reference superpixel 
	\includegraphics[width=\linewidth]{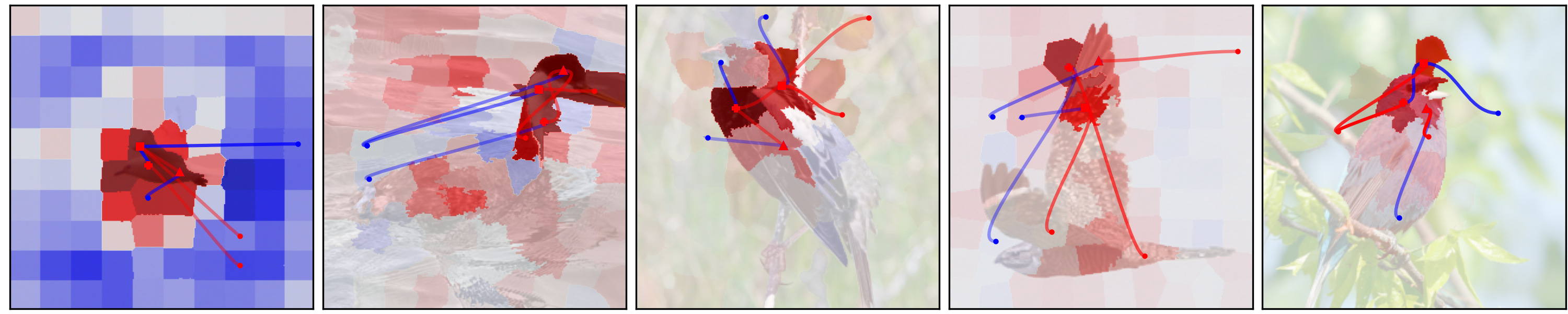}
	(b) Random reference superpixel
	\includegraphics[width=\linewidth]{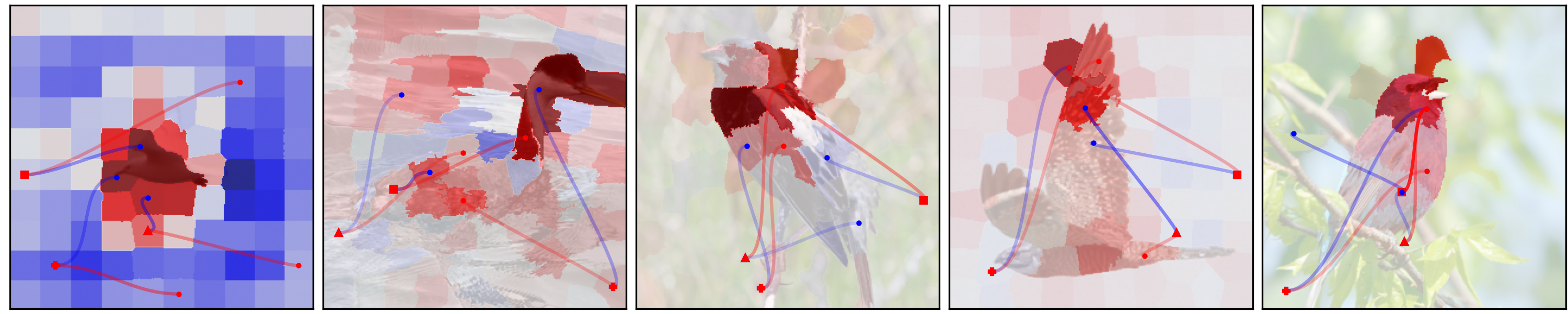}
	\caption{Interaction for the CUB birds dataset for 5 randomly selected samples obtained using a (marginal) train set imputer. The transparency of the red(blue) b\'ezier curves represents the strength of the most positive (negative) raw joint effects. The \pd{} relevances are visualized as heatmaps.
	In the upper panel the three reference superpixel are chosen according to highest relevance. In contrast the lower panel shows random reference superpixels. The predicted class probabilities of the five samples are 0.15, 0.62, 0.89, 0.856 and 0.98.
	}
	\label{fig:cub_train}
\end{figure}

We show attributions for a train set imputer in \Cref{fig:cub_train}. We used the same number of imputations ($\#=100$) as for the results based on the histogram imputer in \Cref{fig:cub}.
While \pd{} relevance attributions are qualitatively similar between the two imputers, the relevance is less concentrated on the central object for the train set imputer. The highest interaction effects with the most relevant superpixels as reference points are mostly similar between train set and histogram imputer. Importantly, the observation that interaction between random superpixels is small can be confirmed for the train set imputer.

\begin{figure}
	\centering
	\includegraphics[clip, trim=0.cm 0.5cm 0.cm 0cm, width=0.7\linewidth]{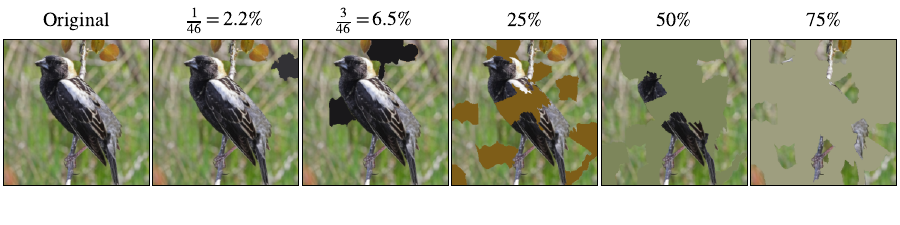}
	\includegraphics[clip, trim=0.cm 0.5cm 0.cm 0cm, width=0.7\linewidth]{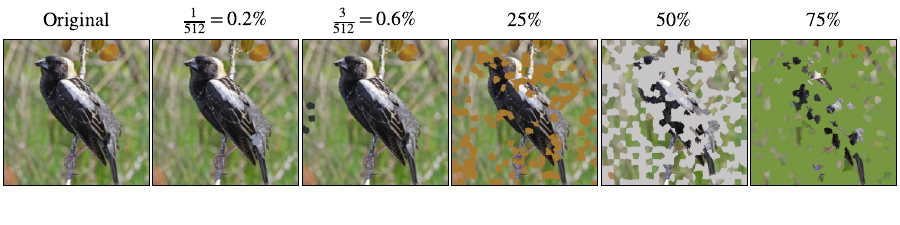}
	\caption{Imputed samples generated with the (conditional) histogram imputer based on 50 (top) and 500 (bottom) SLIC superpixels. Increasingly large fraction is imputed, starting with one, three superpixels up to 75\% of all superpixels. 
			\pd{} only requires a few imputed superpixels. In contrast, a typical Shapley coalition $S$ covers 50\% of all features, which practically leads to increasingly off-manifold samples. 
			}
	\label{fig:cub_imputer}
\end{figure}

Finally, in \Cref{fig:cub_imputer} we visualize the practical challenges for conditional imputers arising from imputing a large fraction of superpixels. This is supposed to support the argument of potential off-manifold model evaluations in these cases.

\printcredits
\section*{Acknowledgements}
This work was funded by the German Ministry for Education and Research as BIFOLD - Berlin Institute for the Foundations of Learning and Data (ref. 01IS18025A and ref. 01IS18037A).

\bibliographystyle{cas-model2-names}

\bibliography{bibfile}

\end{document}